\documentclass[11pt,a4paper]{article}

\usepackage[utf8]{inputenc}
\usepackage[T1]{fontenc}
\usepackage{lmodern}        
\usepackage{geometry}       
\usepackage{setspace}       
\usepackage{graphicx}       
\usepackage{amsmath, amssymb} 
\usepackage{hyperref}       
\usepackage[numbers, sort&compress]{natbib}
\usepackage{svg}

\usepackage{algorithm}
\usepackage{algpseudocode}
\usepackage{amsmath}
\usepackage{graphicx} 
\usepackage{subcaption} 
\usepackage{booktabs}  
\usepackage{multirow} 
\usepackage{tikz}
\usetikzlibrary{arrows.meta, positioning}

\usepackage{enumitem}
\usepackage[blocks]{authblk} 

\usepackage{float}

\begin{document}

\title{The Simulacrum: Decision‑Theoretic Pretraining for Near‑Optimal Time‑Series Forecasting and Inference
\thanks{Preliminary versions of the core ideas in this paper were presented in earlier conference and invited talks, including CMStatistics 2022 \citep{montero2022big}, the International Symposium on Forecasting 2023\citep{montero2023pretrained} , and the M6 Competition Conference 2023 \citep{montero2023m6}. The present paper provides the first unified decision-theoretic formulation of this framework, together with comprehensive empirical validation. }}

\author[]{Pablo Montero-Manso\thanks{Part of this work was developed while the author was a Visiting Researcher at Google. The views expressed are those of the authors and do not necessarily reflect those of Google.} }
\author[]{Marcel Scharth}

\affil[]{University of Sydney Business School, Australia}

\date{Work in Progress: \today}

\maketitle

\begin{abstract}
We introduce a neural network-based framework for learning time series estimators through a process we term \textit{decision-theoretic pretraining}.
Analysts specify a \textit{generative world}, a distribution over data-generating processes, and a target \textit{decision objective}.
A neural network trained on stratified simulations from this world approximates the corresponding optimal decision rule, yielding a \textit{neural estimator} that provides forecasts, parameter estimates, predictive intervals, or model-selection for zero-shot inference on previously unseen time series.

The joint specification of the generative world and objective enables the estimators to directly approximate process-level, finite-sample properties: near-optimal risk, bias control, minimax performance, and uniform calibration.
Our experiments demonstrate that these neural estimators can outperform traditional baselines such as maximum likelihood estimation and model selection via AICc, for the same model structural model classes.
Furthermore, even when trained purely on simulations of structural models, they achieve competitive or state-of-the-art forecasting accuracy on major real-world benchmarks, compared with statistical, neural or large pre-trained models.

We illustrate the framework by addressing two longstanding challenges: finite-sample bias and miscalibration in AR($p$) models, and the forecast combination puzzle.
These applications highlight the approach's main advantage: its ability to approximate solutions to analytically intractable or computationally prohibitive time series problems, including complex structural equations or optimality criteria.
Ultimately, by enabling explicit control over decision-theoretic trade-offs, the framework equips analysts with highly efficient estimation tools tailored to their specific analytical needs.
\end{abstract}


\section{Introduction }

Time series models are a central analytical tool across science and industry,
used to infer the properties of evolving processes and forecast their trajectories.
Traditionally, researchers rely on structural models:
frameworks that impose specific mathematical assumptions to capture a system's dynamics or decompose it
into interpretable components such as trends, cycles, persistence, and regime changes.
Although parsimonious and intentionally stylized, structural models are often sufficiently expressive for practical use.
Their usefulness, however, depends not only on whether the assumed structure is appropriate,
but on whether its parameters, states, and predictive implications can be estimated reliably.
As we discuss in Section \ref{subsec:motivation}, the main limitation of structural models is frequently not the structure itself, but how the parameters
are estimated from data.
Well-known estimators for popular models still suffer from data inefficiency,
finite-sample bias, numerical instability, and miscalibration.
These flaws propagate into the model's predictions; consequently, improvements in estimation directly yield better forecast accuracy and more reliable inference.

Recently, deep learning forecasting methods, including Time Series Foundation Models (TSFMs) \citep{ansari2024chronos, das2024timesfm} and Global Forecasting Models (GFMs) \citep{salinas2020deepar, oreshkin2020nbeats, lim2021temporal},
have achieved strong empirical accuracy by training flexible neural architectures on large, diverse collections of time series.
Their standard training objectives, however, are not designed to deliver process-level,
finite-sample guarantees or parsimonious structural interpretation.
Parsimonious assumptions are not merely a route to better forecasts; they are often the primary object of interest,
allowing researchers to test the validity of a mechanism, maintain interpretability,
and ensure auditable results. Furthermore, statistical guarantees at the individual-series level (such as risk coverage for a specific asset within a portfolio) are difficult to achieve with TSFMs and GFMs.
Because these models optimize for average performance across heterogeneous datasets,
strong aggregate metrics can hide systematic errors in particular series or parameter regimes.
This issue cannot be solved by pure empirical data scaling: even within massive datasets,
the history of any single series remains a ``small sample'' lacking sufficient information to offer formal guarantees.

To bridge this gap, this paper introduces the \emph{Simulacrum},
a computational framework that recasts time series estimation as a statistical decision problem \citep{wald1950statistical, berger2013statistical} and solves it by simulation. The analyst specifies two elements: a \emph{generative world}, a distribution over data-generating processes on which the estimator should perform well, and a \emph{decision objective} that encodes the desired notion of performance. Together these define an optimal decision rule. Such a rule is rarely available by classical means: maximum likelihood is optimal only asymptotically and biased in finite samples \citep{marriott1954bias}, while minimax estimators are well defined but have no closed form. 
Through a process we term \emph{decision-theoretic pretraining}, we utilize massive-scale simulations to train neural networks to approximate this optimal rule directly.
By doing so, we enable the derivation of near-optimal \textit{neural estimators} for both existing and novel model classes.

\begin{figure}[!htbp]
  \centering
  \includegraphics[width=\linewidth]{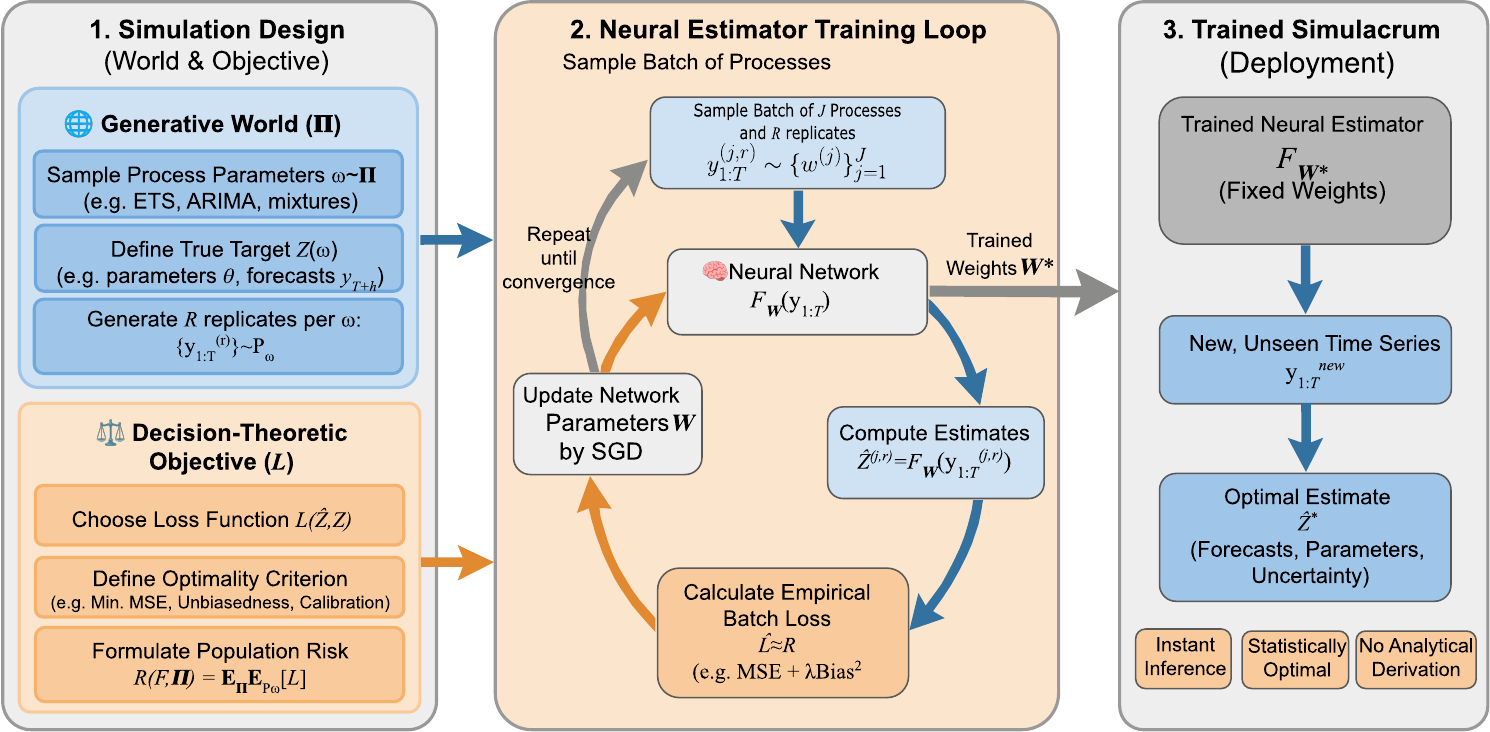}
  \caption{\textbf{Overview of the Simulacrum framework.} The analyst
  specifies a generative world $\Pi$, which defines the mechanisms,
  parameter ranges, and possible contaminations on which the estimator
  should perform well. Training uses a two-stage simulation scheme: an
  outer loop samples data-generating processes from the world, and an
  inner loop generates replicated trajectories from each process. These
  simulations provide supervised targets and process-level Monte Carlo
  estimates, allowing a neural network to learn near-optimal forecasting
  or inference rules.}
  \label{fig:simulacrum}
\end{figure}

The framework’s core innovation is a two-stage, stratified simulation scheme. An outer loop samples model parameters within a general model class (the \emph{generative world}), while an inner loop generates replicate trajectories from that specific set of parameters. A world can include multiple models, contamination operators, and data augmentations. With enough replicates, a loss function can measure the neural estimator's deviations from process-level optimality (e.g., conditional bias or miscalibration). The network is trained to minimize this loss, converging to an approximation of the optimal decision rule for the entire model class. For example, consider estimating a short AR($5$) series. Standard estimators such as maximum likelihood are biased in finite samples, and their resulting intervals are rarely uniformly calibrated. Our framework samples a world of stationary AR($5$) coefficient vectors, generates multiple independent trajectories from each, and trains a neural network to recover coefficients or predictive quantiles from a single observed trajectory. The outer loop supplies the coefficient vectors, while the inner loop provides the replicates that make conditional bias or miscoverage measurable and penalizable in training, process by process. The trained network becomes a fast, finite-sample estimator whose statistical properties are deliberately fixed by the chosen world and objective. See Figure~\ref{fig:simulacrum} for an overview of the framework.

The world perspective situates the framework relative to recent work on neural forecasting. Global forecasting models, time series foundation models, and prior-data fitted networks \citep{muller2022transformers} can each be viewed as training an estimator over a generative world, such as an empirical corpus, a large pretraining mixture, or a synthetic prior. In these approaches, however, the objective is typically fixed: minimizing risk over the world. Our contribution is to make the estimator itself an object of design. The analyst specifies the world together with the properties the resulting estimator should satisfy. When the world is simulated, we can target and verify these properties directly. In the special case of a single Bayesian model under average risk, the framework coincides with neural Bayes estimation \citep{SZH24}; its reach is the generalization beyond that case, to objectives and model classes whose optimal estimators were previously unavailable.

Our experiments showcase the world design principles: trained for accuracy, neural estimators outperform maximum likelihood under the exponential smoothing and AR($p$) worlds; trained for bias control, they achieve near-unbiased finite-sample estimation where maximum likelihood is systematically biased; trained for worst-case risk, they flatten the risk profile and reduce the largest forecast errors. Two further results address long standing difficulties in classical estimation. For the AR($p$) class, the framework yields estimation intervals that come much closer to uniform calibration (coverage holding conditional on each parameter value) than classical plug-in procedures or standard empirical risk minimization, which exhibit systematic conditional miscalibration. For forecast combination, we recast the heuristic of averaging forecasts as estimation within a generative class: observations are modeled as a convex combination of ARIMA and exponential smoothing components driven by common shocks, a class for which the optimal estimator is analytically intractable. Learning this combination rule directly converts a long-standing heuristic into a principled method and yields a generative account of the Forecast Combination Puzzle \citep{clemen1989combining, smith2009simple}.

These design choices also translate into competitive accuracy on real data. Although the neural estimators are trained entirely on simulated structural worlds and never see the test series, they are competitive with, and often surpass, both classical and modern alternatives on standard benchmarks. On the M1 competition data, a neural estimator for the additive exponential smoothing class attains the lowest error among all 21 competition entrants on the competition's median error metric, including several alternative estimators for the exponential smoothing class. On the Monash archive, the combined ARIMA--ETS estimator improves on the strongest reported baselines for several datasets.

\subsection*{Contributions and Paper Organization}
\label{subsec:contributions}

The paper makes three main contributions.

\begin{itemize}

    \item \textbf{A unified simulation-based framework for structural time-series estimation (Sections~\ref{sec:framework}--\ref{sec:neural_exp}):}
    We introduce \emph{The Simulacrum}, a decision-theoretic framework that trains neural networks as estimators over user-defined generative worlds. The framework extends standard simulation-based training beyond average-risk minimization and makes it possible to target finite-sample properties such as minimax risk, breakdown-point robustness, and uniform calibration. In \S~\ref{sec:framework}, we define generative worlds as a generalization of time-series data-generating processes (DGPs), together with loss functions that encode statistical objectives. Section~\ref{sec:simulation-based-training} develops the two-stage simulation mechanism used to train neural estimators, and Section~\ref{sec:neural_exp} discusses neural architectural considerations. Connections to neighboring paradigms such as amortized inference, Prior-Data Fitted Networks (PFNs), and global or time series foundation models are discussed later in \S~\ref{sec:related-work}.

    \item \textbf{Methodological results on finite-sample estimation and uncertainty quantification (Sections~\ref{sec:ETS}--\ref{sec:ARp}):}
    The framework recovers estimators with statistical properties that are difficult to obtain with classical procedures. In the Exponential Smoothing (ETS) class, neural estimators trained for forecasting accuracy achieve superior performance relative to standard alternatives such as Maximum Likelihood, both under the true DGP (\S~\ref{sec:ETS_def_inference}) and in real data (\S~\ref{sec:ETS_empiric}, compared with five ETS estimators and 21 competing models on the M1 dataset). In the AR($p$) class (Section~\ref{sec:ARp}), the framework yields estimators with near-unbiased finite-sample behavior and \textit{uniform calibration}, a stronger property than the marginal calibration typically achieved by standard deep learning approaches. These results show that simulation-based neural estimators can be trained to satisfy process-level criteria, not only aggregate predictive accuracy.

    \item \textbf{Empirical demonstrations in model selection, forecast combination, and comparative risk analysis (Section~\ref{sec:modelselpuzzle} and related results):}
    We further show that the same framework applies naturally to multi-model worlds. In Section~\ref{sec:modelselpuzzle}, we propose a generative treatment of the ``Forecast Combination Puzzle'' by modeling forecast combination itself as a structural process. This yields direct neural estimators of combination rules without relying on the standard two-step procedure of first fitting component models and then estimating weights. In real data, the resulting estimator is competitive with, and in some cases improves upon, simple equal-weight combinations, while also clarifying why equal weights remain a strong benchmark. More broadly, the paper includes comparative risk and scaling analyses that characterize when neural estimators outperform classical alternatives and when the gap narrows. In \S~\ref{sec:ETS_scaling}, we study how approximation rates vary with series length and model complexity. We also compare neural \textit{model selection} for ETS subclasses with AIC-based selection as series length increases (\S~\ref{sec:ETS_modelsel}), and provide detailed error curves against established baselines, including Maximum Likelihood and Bayesian MCMC for ETS (\S~\ref{sec:ETS_def_inference}), as well as OLS, Yule--Walker, and Burg for the AR($p$) class (\S~\ref{sec:arpriskbias}).

\end{itemize}

\paragraph{Reading Guide.}
Readers primarily interested in practical forecasting performance and applied tools may wish to begin with Section~\ref{sec:ETS_empiric} and Section~\ref{sec:combipuzzle_results}, which present real-data results for exponential smoothing and forecast combination on benchmark datasets. Sections~\ref{sec:framework}--\ref{sec:simulation-based-training} can then be read as providing a unifying explanation of how these applications arise from a common training principle.

Readers interested in statistical foundations, decision theory, or simulation-based inference may prefer to begin with Sections~\ref{sec:framework}--\ref{sec:simulation-based-training}. These sections introduce generative worlds, population risk, and conditional performance training, which form the conceptual backbone of the paper.

Readers from a machine learning perspective may find it useful to view Sections~\ref{sec:framework}--\ref{sec:simulation-based-training} as defining a task distribution, training objectives, and a learning procedure, and Sections~\ref{sec:ETS}--\ref{sec:modelselpuzzle} as applications of this procedure under different world designs and losses.

\section{A Decision Framework for Time Series Estimation}
\label{sec:framework}

This section formulates time series estimation and forecasting as a statistical decision problem \citep{wald1950statistical, berger2013statistical}. The problem is defined by two design choices: a training environment and an objective. We represent the training environment by a \emph{generative world}, a distribution over data-generating processes on which an estimator is trained and evaluated. This plays a role analogous to a task distribution in meta-learning or an environment in reinforcement learning \citep{baxter2000model, finn2017model}. A world specifies which time series processes are included, how their configurations vary, and which forms of noise or contamination are treated as relevant.

Given a world, the analyst defines a loss function that encodes the desired notion of performance for forecasts, estimates, and inferential objects, such as accuracy, calibration, bias control, or worst-case behavior. Together, the world and the loss define a population-level learning objective with a well-defined optimal decision rule. By training neural networks on simulated data drawn from the world, we approximate this optimal rule directly, amortizing the solution to the underlying decision problem; see Section~\ref{sec:simulation-based-training}. The remainder of this section formalizes these components and shows how they combine into a unified simulation-based framework for time series forecasting and inference.

\subsection{Generative Time Series Worlds}

We define a generative world $\Pi$ as a probability distribution over fully
instantiated data generating processes (DGPs),
\[
    \omega = (m,\theta) \sim \Pi,
\]
where $m$ denotes a mechanism, a complete procedure for simulating time series data, and $\theta \in \Theta_m$ is its realized parameter vector.

A mechanism may correspond to a classical parametric model class such as ARIMA,
ETS, or GARCH, but it may also represent model combinations or an empirical resampling scheme. Each $\omega$ induces a distribution $P_\omega$ over observable trajectories and, when present, latent states:
\[
    (Y_{1:T}, S_{1:T}) \sim P_{\omega}.
\]
In practice, this framework can accommodate diverse time series lengths within the world; we suppress this detail in the notation for simplicity.

It is often useful to view $\Pi$ as inducing a factorization
\[
    \Pi = \Pi_m\,\Pi_{\theta\mid m},
\]
which makes explicit that a world specifies two components: a distribution $\Pi_m$ over mechanisms and an instantiation rule $\Pi_{\theta\mid m}$ for each mechanism.

\paragraph{Single-mechanism worlds.}
If the world selects a fixed mechanism $m_\star$ (e.g., a simple exponential smoothing model),
\[
    \omega = (m_\star,\theta),
    \qquad
    \theta \sim \Pi_{\theta},
\]
then the variability across $\omega \sim \Pi$ derives solely from parameter variation. We recover a Bayesian model as the special case where $m_\star$ is a likelihood family and $\Pi_{\theta}$ coincides with a Bayesian prior.

\paragraph{Multi-mechanism worlds.}
When the world allows heterogeneous mechanisms, it first samples
\[
    m \sim \Pi_m,
    \qquad
    \theta \sim \Pi_{\theta\mid m}.
\]
The support of $\Pi$ is then the disjoint union
\[
    \Omega
    =
    \bigsqcup_m
    \bigl(
        \{m\}
        \times
        \Theta_m
    \bigr),
\]
allowing for worlds that include diverse model classes (e.g., ARIMA, exponential smoothing, and GARCH processes) and other types of mechanisms. Bayesian model averaging \citep{hoeting1999bayesian} is a special case in which each $m$ is a Bayesian model with a well-defined likelihood and marginal likelihood, $\Pi_m$ is the model prior, and $\Pi_{\theta\mid m}$ is the corresponding parameter prior.

\subsection{World Design}

The construction of $\Pi$ is an instance of environment design \citep{dennis2020emergent}: specifying which mechanisms, parameter ranges, structural variations, perturbations, and data sources the estimator should be exposed to during training. Training a predictor on a deliberately broad distribution of simulated environments so that it transfers to reality is known as domain randomization \citep{tobin2017domain}; world design applies the same principle to time series estimation.

\paragraph{Synthetic worlds.}
Analytically defined mechanism families, such as ARIMA, ETS, or GARCH, paired with parameter rules $\Pi_{\theta\mid m}$. These worlds offer full control and broad structural coverage.

\paragraph{Empirical worlds.}
An empirical world is formed from an observed collection of time series, using resampling or rolling windows to generate trajectories. 

\paragraph{Hybrid worlds.}
Hybrid worlds combine synthetic and empirical components to balance mechanism coverage with realism. They are closely related to the pretraining distributions used in modern time series foundation models.

\bigskip
This paper focuses on synthetic worlds, which enable a richer class of estimators by making the relevant supervisory quantities observable.

\subsection{Estimation Targets and Supervisory Signals}
\label{sec:targets}

For each process $\omega$ and trajectory $y_{1:T}$, the analyst specifies an
\emph{estimand}
\[
    Z(\omega,y_{1:T}) \in \mathcal Z,
\]
representing the object to be estimated. Depending on the task, this may be a
parameter, a latent state functional, a future value, a predictive functional,
or an entire conditional distribution. The estimator is a map
\[
    F:\mathcal Y_T \to \mathcal Z,
    \qquad
    \widehat Z = F(y_{1:T}),
\]
where $\mathcal Y_T$ denotes the space of observed histories and $\mathcal Z$
denotes the output space appropriate to the estimand.

Examples include:
\begin{itemize}
    \item \textbf{Parameter estimation:} a fixed functional of the DGP parameter,
    \[
        Z(\omega,y_{1:T}) = g(\theta),
    \]
    often simply $Z(\omega)=\theta$.

    \item \textbf{Point forecasting:} a functional of the predictive distribution,
    such as the conditional mean
    \[
        Z(\omega,y_{1:T})
        =
        \mathbb E_\omega(Y_{T+h}\mid y_{1:T}).
    \]

    \item \textbf{Quantile forecasting:} a conditional predictive quantile,
    \[
        Z(\omega,y_{1:T})
        =
        q^\omega_\alpha(Y_{T+h}\mid y_{1:T}),
    \]
    where $q^\omega_\alpha$ denotes the conditional $\alpha$-quantile.

    \item \textbf{State estimation:} a conditional functional of latent states,
    such as
    \[
        Z(\omega,y_{1:T})
        =
        \mathbb E_\omega(S_{1:T}\mid y_{1:T}).
    \]

    \item \textbf{Distributional outputs:} an entire conditional object, such as
    a predictive distribution,
    \[
        Z(\omega,y_{1:T})
        =
        P_\omega(\cdot\mid y_{1:T}).
    \]
\end{itemize}

In synthetic worlds, the simulator often supplies a realized quantity that can
be used to train an estimator for $Z$. We denote this realized supervisory
signal by $\widetilde Z$. For some targets, the estimand is directly observed by construction. For example, in parameter estimation the simulator draws $\theta$, so for a target $Z(\omega, y_{1:T})=g(\theta)$ we can set
\[
    \widetilde Z = g(\theta) 
\]
For other targets, the estimand is a conditional functional and is not usually
available in closed form. In this case, we choose $\widetilde Z$ so that minimizing the expected loss against $\widetilde Z$ yields $Z$ as the population target \citep{gneiting2011making}:
\[
    Z(\omega,y_{1:T})
    \in
    \arg\min_{z\in\mathcal Z}
    \mathbb E_\omega
    \left[
        L(z,\widetilde Z)
        \mid
        Y_{1:T}=y_{1:T}
    \right].
\]

\subsection{Loss Functions and Decision Rules}

We perform evaluation through a user-specified loss function
\[
    L\bigl(F(y_{1:T}),\,\widetilde Z\bigr),
\]
which may represent squared error, absolute error, pinball loss, or another objective.

Given a world $\Pi$, we define the process-level risk of a decision rule $F$
under a DGP $\omega$ as
\[
    r(\omega,F)
    =
    \mathbb{E}_{Y_{1:T},\,\widetilde Z\sim P_\omega}
    \left[
        L\bigl(F(Y_{1:T}),\,\widetilde Z\bigr)
    \right],
\]
where the expectation is over the observed history and, where applicable, the supervisory draw conditional on that history. The minimizer $F^\star$ targets the loss-implied estimand $Z$ defined in Section~\ref{sec:targets}. 

The \emph{population risk} of a decision rule $F$ is
\begin{equation}
\label{eq:risk}
    \mathcal R(F;\Pi)
    =
    \mathbb{E}_{\omega\sim\Pi}
    \bigl[
        r(\omega,F)
    \bigr].
\end{equation}
The minimizer
\[
    F^\star \in \arg\min_F \mathcal R(F;\Pi)
\]
is the optimal decision rule under that world. The function $F^\star$ represents the ideal estimator or forecasting rule for the design choices encoded by $(\Pi,L)$. When $\Pi$ corresponds to a Bayesian model and the loss is separable, the population risk \eqref{eq:risk} becomes the prior predictive risk. Its pointwise minimizer is the usual Bayes action under the posterior predictive distribution.

\subsection{Process-Level Properties}
\label{sec:restrictions}

In many applications, minimizing the population risk $\mathcal R(F;\Pi)$ is insufficient. The analyst often requires the decision rule $F$ to satisfy statistical properties such as bias control or calibration across the support of the world.

We focus on targeting uniform properties, defined as constraints on the distribution of $F(Y_{1:T})$ at each fixed DGP $\omega$. In contrast, marginal properties constrain only the unconditional distribution induced by the mixture over $\Pi$.

Formally, the uniform problem is
\begin{equation}
\label{eq:constrained-problem}
    \min_F \; \mathcal R(F;\Pi)
    \qquad
    \text{s.t.}
    \qquad
    C_k(\omega,F) \le \delta_k
    \quad
    \forall\,\omega\in\operatorname{supp}(\Pi),
    \quad
    k=1,\dots,K,
\end{equation}
where each $C_k(\omega,F)$ is a process-level property of interest.

Two important examples are:
\begin{itemize}
    \item \emph{Bias control.}
    For process-level estimands $Z(\omega)$, such as fixed parameters or other DGP-level functionals,
    \[
        C_{\mathrm{bias}}(\omega,F)
        =
        \left\|
            \mathbb E_\omega[F(Y_{1:T})]
            -
            Z(\omega)
        \right\|.
    \]

    \item \emph{Process-level calibration.}
    For a predictive quantile estimator $q_\alpha(y_{1:T})$ at nominal coverage level $\alpha\in(0,1)$,
    \[
        C_{\mathrm{cal}}(\omega,F)
        =
        \left|
            \mathbb P_\omega
            \left\{
                Y_{T+h} \le q_\alpha(Y_{1:T})
            \right\}
            -
            \alpha
        \right|.
    \]
\end{itemize}

Direct optimization of \eqref{eq:constrained-problem} is intractable since each $C_k(\omega,F)$ is itself a population-level functional of the distribution under $\omega$. The training procedures of Section~\ref{sec:simulation-based-training} approximate these constraints through differentiable penalties computed from replicated simulations. Depending on the application, these penalties can target average violations under $\Pi$, stratified violations over regions of the world, or smooth approximations to worst-case violations.

\subsection{Worlds vs. Models}
\label{sec:worlds-decision-environments}

A world $\Pi$ specifies the range of processes over which a decision rule should perform well. Although any world can often be embedded in a sufficiently rich hierarchical probability model, this embedding obscures the role that $\Pi$ plays in the present framework. 

A traditional model-based workflow typically fixes a probability family $P_\theta$ and uses observed data to infer $\theta$ or functions thereof. In our framework, the primary object is instead the estimator itself:
\[
    F:\mathcal Y \to \mathcal Z.
\]
The world $\Pi$, loss $L$, and constraint functionals $C_k$ completely specify the desired properties of this estimator. Simulation-based training then approximates an optimal decision rule satisfying this specification. In this sense, the procedure amortizes an inference machine \citep{gershman2014amortized}. 

This shift is why we refer to $\Pi$ as a world rather than a model. Modifying $\Pi$ engineers the competence profile of the resulting inference machine: dictating which mechanisms it recognizes, which regimes it handles gracefully, which data pathologies it is robust against, and which statistical trade-offs it internalizes.

\subsection{Robust Design}
\label{sec:robust-design}

Real-world time series frequently exhibit irregularities such as outliers, missing values, level shifts, censoring, and measurement noise.  To encode robustness at the level of the world, we introduce contamination variables and corruption operators, following the contamination-neighborhood formulation of robust statistics \citep{huber1992robust}.

Let $\omega_{\mathrm{clean}}=(m,\theta)$ denote a clean process and let $\xi$ denote a contamination specification drawn from a distribution $\Pi_\xi$. A robust world draws
\[
    \omega_{\mathrm{clean}} \sim \Pi_{\mathrm{clean}},
    \qquad
    \xi \sim \Pi_{\xi\mid \omega_{\mathrm{clean}}},
\]
and generates data by applying a corruption operator $C_\xi$ to the clean process:
\[
    (Y_{1:T},S_{1:T}) \sim P_{\omega},
    \qquad
    P_\omega = C_\xi \# P_{\omega_{\mathrm{clean}}},
\]
where $C_\xi \# P_{\omega_{\mathrm{clean}}}$ denotes the distribution induced by applying $C_\xi$ to draws from the clean process. Here $C_\xi$ may encode additive outliers, missingness masks, censoring rules, level shifts, measurement noise, or other perturbations.

The robust world can therefore be regarded as a distribution over contaminated processes $\omega=(\omega_{\mathrm{clean}},\xi)$, even though the clean generative world is written in the simpler form $\omega=(m,\theta)$. Specifying contamination as part of the world design allows the estimator to learn structural responses to irregularities rather than treating them only as post hoc exceptions.

This completes the definition of the unified decision framework. For the remainder of the paper, we treat $\Pi$ as fixed and study the decision rules that arise from minimizing \eqref{eq:risk} or its generalizations.

\section{Simulation-Based Training}
\label{sec:simulation-based-training}

Section~\ref{sec:framework} defined the target decision rule as the optimizer of a population objective over a generative world. We now describe how we can approximate this objective in practice by simulation.

\subsection{Training algorithm}
\label{sec:training_algorithm}

We train the neural estimator $F$ by stochastic gradient descent (SGD) on a Monte Carlo approximation of the population risk \eqref{eq:risk}. We construct each mini-batch in two stages: an outer loop samples $J$ data generating processes from $\Pi$, and an inner loop generates $R$ independent trajectories from each sampled process. The total batch size is $B=JR$.

Formally, at each SGD update we:
\begin{enumerate}
    \item Sample $J$ processes from $\Pi$:
    \[
        \omega^{(j)}
        =
        (m^{(j)},\theta^{(j)})
        \sim \Pi,
        \qquad
        j=1,\dots,J.
    \]

    \item For each $\omega^{(j)}$, sample $R$ i.i.d.\ trajectories:
    \[
        (Y^{(j,r)}_{1:T}, S^{(j,r)}_{1:T})
        \overset{\mathrm{i.i.d.}}{\sim}
        P_{\omega^{(j)}},
        \qquad
        r=1,\dots,R.
    \]

    \item For each replicate, compute the network output
    \[
        F^{(j,r)}
        =
        F(y^{(j,r)}_{1:T})
    \]
    and the supervisory signal $\widetilde Z^{(j,r)}$. For parameter targets,
    $\widetilde Z^{(j,r)}=g(\theta^{(j)})$, often simply $\theta^{(j)}$ itself. For latent-state targets,
    $\widetilde Z^{(j,r)}=S^{(j,r)}_{1:T}$. For predictive targets,
    $\widetilde Z^{(j,r)}$ is the realized future value or path, such as
    $Y^{(j,r)}_{T+h}$ or $Y^{(j,r)}_{T+1:T+H}$. 

    \item Form the empirical batch loss $\widehat L$ from these quantities.

    \item Update $F$ by stochastic gradient descent on $\widehat L$.
\end{enumerate}

The simplest case is empirical risk minimization,
\[
    \widehat L_{\mathrm{ERM}}(F)
    =
    \frac{1}{JR}
    \sum_{j=1}^J
    \sum_{r=1}^{R}
    L\bigl(
        F(y^{(j,r)}_{1:T}),
        \widetilde Z^{(j,r)}
    \bigr).
\]
For ordinary risk minimization, $R=1$ is sufficient. Having $R>1$ replicates from a fixed DGP enables Monte Carlo estimation of process-level functionals of $F$ at that DGP. This is what allows the framework to target the properties introduced in Section~\ref{sec:restrictions}. The next subsections construct losses for bias control, calibration, and minimax estimation. 

\subsection{Targeting Process-Level Properties}
\label{sec:frequentist_properties}

The constrained problem in \eqref{eq:constrained-problem} is an ideal population formulation. In practice, we approximate it through differentiable penalties computed from replicated simulations. The key idea is that the outer loop samples processes $\omega^{(j)}$, while the inner loop provides repeated histories from each fixed process. These repeated histories allow us to estimate properties of $F$ conditional on the DGP, such as process-level bias or coverage.

\subsection{Bias control}
\label{sec:bias-reduction}

Classical time series estimators are often biased in finite samples, and bias-correction procedures, when available, are usually model-specific: analytic bias expansions must be derived separately for each model class \citep{marriott1954bias, shaman1988bias}, and bootstrap-based corrections are constructed in a similarly bespoke way \citep{kilian1998small}. The simulation framework provides a more general route: by generating replicated histories at fixed $\omega$, we can estimate process-level properties of the decision rule $F$ directly.

We begin with bias control for estimands of the form $Z(\omega)$, such as parameters or other fixed functionals of the DGP, that do not depend on the realized trajectory $y_{1:T}$. For these targets, the supervisory signal is deterministic conditional on $\omega$, so $\widetilde Z=Z(\omega)$. The bias of $F$ at process $\omega$ is
\[
    b(\omega)
    =
    \mathbb E_\omega
    \left[
        F(Y_{1:T})
    \right]
    -
    Z(\omega).
\]
Given $R$ replicated trajectories from $\omega^{(j)}$, we estimate this bias by
\[
    \widehat b(\omega^{(j)})
    =
    \frac{1}{R}
    \sum_{r=1}^R
    F\bigl(y^{(j,r)}_{1:T}\bigr)
    -
    Z(\omega^{(j)}).
\]

There are several ways to aggregate these process-level estimates across sampled DGPs. The ideal uniform constraint controls the worst process in the support of $\Pi$, but in finite training we observe only a sampled set of processes. In our experiments, we use the average squared bias over sampled processes as a relaxation. A base estimator $F_0$ is first trained to minimize $\widehat L_{\mathrm{ERM}}(F)$. A second training phase then optimizes
\[
    \widehat L_{\mathrm{bias}}(F)
    =
    \widehat L_{\mathrm{ERM}}(F)
    +
    \lambda_b
    \frac{1}{J}
    \sum_{j=1}^J
    \|\widehat b(\omega^{(j)})\|^2 .
\]
Training proceeds along a short path of penalty weights $\lambda_b$, with each stage warm-started from the previous one. The final estimator is selected to balance empirical risk against squared bias. This objective is a Lagrangian relaxation of
\[
    \min_F \mathcal R(F;\Pi)
    \qquad
    \text{subject to}
    \qquad
    \mathbb E_{\omega\sim\Pi}
    \left[
        \|b(\omega)\|^2
    \right]
    \le
    \delta^2,
\]
with $\lambda_b$ acting as the dual variable.

This approach encourages low bias across the sampled world, but is not a uniform constraint. Stronger sampled-uniform variants can be obtained by replacing the average over $j$ with log-sum-exp aggregation as in Section~\ref{subsec:minimaxtrain}. When the outer sampling scheme gives adequate coverage of the relevant regions of $\Pi$, the average penalty can nevertheless provide effective empirical control across the world while remaining more stable to optimize than a worst-case objective.

More broadly, process-level Monte Carlo estimates make other forms of bias reduction possible. One possible extension is to train a \emph{bias-correction head}: a sub-module that learns a correction $\mathcal C(y)$ on top of a frozen base estimator $F_0$, yielding
\[
    F(y)
    =
    F_0(y)
    +
    \mathcal C(y).
\]
The correction can be trained using the same replicate-based penalty, so that systematic deviations of $F_0$ from $Z(\omega)$ are reduced across regions of the process space.

\subsection{Calibration and Coverage}
\label{sec:calibration}

Calibration is, together with sharpness, the standard basis for evaluating probabilistic forecasts \citep{Gneiting2007}. Whereas that paradigm treats calibration as a diagnostic applied after estimation, the simulation framework makes it an explicit training target.

Suppose $F$ delivers a predictive quantile
\[
    q_\alpha(y_{1:T})
\]
for a scalar realized target $\widetilde Z$, such as $\widetilde Z=Y_{T+h}$, at nominal coverage level $\alpha\in(0,1)$. The desired coverage property is
\[
    \mathbb P_\omega
    \left\{
        \widetilde Z
        \le
        q_\alpha(Y_{1:T})
    \right\}
    \approx
    \alpha.
\]
This is a process-level coverage target, required to hold at each fixed $\omega$. It is therefore stronger, but more model-dependent, than the marginal, distribution-free coverage of conformal prediction \citep{vovk2005algorithmic} and its variants for non-exchangeable time series \citep{gibbs2021adaptive}. 

Using $R$ replicates for $\omega^{(j)}$, we approximate process-level coverage by
\begin{equation}
\label{eq:cov_hat_main}
    \widehat{\mathrm{cov}}_{\alpha}
    \bigl(\omega^{(j)}\bigr)
    =
    \frac{1}{R}
    \sum_{r=1}^R
    \mathbf 1
    \left\{
        \widetilde Z^{(j,r)}
        \le
        q_\alpha
        \bigl(
            y^{(j,r)}_{1:T}
        \bigr)
    \right\}.
\end{equation}
A simple calibration loss is
\begin{equation}
\label{eq:loss_calib_main}
    L_{\mathrm{calib}}(F)
    =
    \frac{1}{J}
    \sum_{j=1}^J
    \left(
        \widehat{\mathrm{cov}}_{\alpha}
        \bigl(\omega^{(j)}\bigr)
        -
        \alpha
    \right)^2.
\end{equation}

To obtain gradients, we replace the indicator in \eqref{eq:cov_hat_main} by a smooth surrogate such as
\[
    \sigma
    \left(
        \kappa
        \{
            q_\alpha(y_{1:T})
            -
            \widetilde Z
        \}
    \right),
\]
with slope parameter $\kappa>0$, leaving the overall loss structure unchanged. The slope $\kappa$ controls the bias--variance tradeoff of the surrogate: large values better approximate the indicator but can yield unstable or sparse gradients.

In practice, we combine calibration with a standard quantile, or pinball, loss \cite{koenker1978regression}:
\[
    L_{\mathrm{quantile+calib}}(F)
    =
    L_{\mathrm{pinball}}(F)
    +
    \lambda_{\mathrm c}
    L_{\mathrm{calib}}(F),
\]
and adjust $\lambda_{\mathrm c}$ so that empirical coverage is close to the nominal level. The pinball loss anchors the location of the quantile, while the calibration penalty adjusts process-level coverage.

Extensions to simultaneous prediction bands over a forecast path up to horizon $H$ require specifying whether coverage is marginal at each horizon or joint over the entire path. For joint bands, the coverage event in \eqref{eq:cov_hat_main} is replaced by an event involving the full realized path $\widetilde Z=Y_{T+1:T+H}$, such as inclusion of all future observations within the predicted band; see Appendix~\ref{app:bands}. Training may then use a differentiable surrogate for this joint event, for example a smooth approximation to the maximum violation across horizons.

\subsection{Minimax Training}
\label{subsec:minimaxtrain}

For some applications, we care about worst-case expected loss across the support of $\Pi$:
\[
    \min_F
    \sup_{\omega\in\operatorname{supp}(\Pi)}
    r(\omega,F),
\]
the minimax decision criterion \citep{wald1950statistical}.

Within each training batch, we estimate the process-level risk at each sampled DGP by
\[
    \widehat r_j(F)
    =
    \frac{1}{R}
    \sum_{r=1}^R
    L\bigl(
        F(y^{(j,r)}_{1:T}),
        \widetilde Z^{(j,r)}
    \bigr).
\]
Direct optimization of $\max_j \widehat r_j(F)$ is unstable for SGD because gradients flow through only one process per update. We replace the maximum by its log-sum-exp smoothing:
\begin{equation}
\label{eq:logsumexp}
    L_{\mathrm{minimax}}(F)
    =
    \frac{1}{\psi}
    \log
    \left(
        \sum_{j=1}^J
        \exp
        \left\{
            \psi\,\widehat r_j(F)
        \right\}
    \right),
\end{equation}
where $\psi>0$ is a temperature parameter. This gives the standard smooth approximation
\[
    \max_j \widehat r_j(F)
    \le
    L_{\mathrm{minimax}}(F)
    \le
    \max_j \widehat r_j(F)
    +
    \frac{\log J}{\psi}.
\]
The relaxation approximates the maximum sampled process-level risk at finite $\psi$ and recovers it as $\psi\to\infty$. Gradients flow through all sampled processes with weights proportional to
\[
    \exp\{\psi\,\widehat r_j(F)\},
\]
so the hardest sampled processes dominate as $\psi$ grows. In practice, we choose a moderate $\psi$ and, if necessary, increase it over training to gradually focus on the worst processes in the batch.

This objective is a sampled minimax approximation: it targets the worst process represented in the current batch. When the support of $\Pi$ is large or contains low-probability but high-risk regions, the quality of the approximation depends on the outer sampling scheme. Stratified sampling or adversarial sampling over $\omega$ can be used when more uniform coverage of the world is required.

The same construction extends to any non-negative per-process functional. Replacing $\widehat r_j(F)$ in \eqref{eq:logsumexp} with
\[
    \|\widehat b(\omega^{(j)})\|^2
\]
gives a stronger variant of the bias-control method of Section~\ref{sec:bias-reduction}: the penalty targets the largest sampled squared bias rather than the average squared bias under $\Pi$. Similarly, replacing $\widehat r_j(F)$ with the per-process calibration error
\[
    \left(
        \widehat{\mathrm{cov}}_\alpha(\omega^{(j)})
        -
        \alpha
    \right)^2
\]
targets the largest sampled calibration violation.

\section{Neural Architectures and Experimental Setup}
\label{sec:neural_exp}

The framework described in Sections~\ref{sec:framework}--\ref{sec:simulation-based-training} is agnostic about the particular neural architecture used to approximate the target decision rule. In principle, any sufficiently flexible function approximator could be employed. In practice, however, architecture matters for two reasons. First, it can improve optimization and generalization by introducing inductive biases that match the structure of the estimation problem. Second, it can enforce properties that are known to hold, or are desirable to impose, in a given setting, such as invariances or equivariances with respect to location, scale, trend, or finite temporal context. These considerations are particularly useful when the statistical objective is itself structured, for example when one seeks affine-equivariant forecasts, stable estimation under rescaling, or estimators that depend only on recent observations. A fuller discussion of these architectural considerations is given in Appendix~\ref{sec:invariances}.

In this paper, the architectural choices are deliberately kept simple so that the empirical comparisons focus on the proposed simulation-based training framework rather than on architecture engineering. Unless otherwise stated, the experiments use a feedforward DenseNet backbone \cite{huang2017densely} as a general-purpose function approximator. We explicitly enforce location and scale equivariance through reversible per-series normalization, which reduces the need to simulate over arbitrary ranges of levels and variances and helps align the network with known properties of models such as ARIMA and exponential smoothing. More specialized constraints, such as additive, transformation, or time-translation equivariances, can also be incorporated within the same framework when they are appropriate for the model class or the estimation target; these extensions are described in Appendix~\ref{sec:invariances}.

To accommodate variable-length time series within a common architecture, inputs are padded to a fixed maximum window and the padded values are encoded separately from the normalized data range. This allows the same network to process series of different lengths while keeping the implementation simple. Throughout the applications that follow, we therefore use a shared architectural template and vary the training world, loss function, and target quantity according to the statistical objective of interest. The next sections show that, even with this relatively generic architecture, the proposed training principle is sufficient to deliver substantial gains in forecasting accuracy, finite-sample bias, calibration, and worst-case performance across several model classes.

\section{Application I: Exponential Smoothing}
\label{sec:ETS}

In this section, we illustrate neural estimators using the popular Exponential Smoothing model class as the underlying process.
Exponential Smoothing is a simple model (1 to 3 parameters) that is remarkably accurate in real data.
It is ideal to highlight the impact of neural estimators vs alternative estimators without obfuscating via complex model classes or sophisticated sampling mechanisms.
The three parameters in Exponential smoothing are defined in the range $(0,1)$ so a simulation mechanism that uses the Uniform distribution is parsimonious.
The experiments illustrate four uses of neural estimators:

\begin{itemize}
    \item \textbf{Finite-sample estimation.} Section~\ref{sec:ETS_def_inference} compares neural estimators with maximum likelihood and Bayesian alternatives under the true exponential-smoothing world, focusing on risk, bias, and minimax behavior.
    \item \textbf{Forecasting and scaling.} Section~\ref{sec:ETS_scaling} studies how forecast accuracy changes with horizon, series length, model complexity, and simulation budget.
    \item \textbf{Real-data transfer.} Section~\ref{sec:ETS_empiric} evaluates whether estimators trained only on synthetic exponential-smoothing worlds transfer to real benchmark data from the M1 competition.
    \item \textbf{Tail-risk behavior.} Section~\ref{sec:ETS_minimax} examines whether minimax-oriented training reduces large forecast errors in real data, using the M3 yearly series.
\end{itemize}

Together, these experiments show that the same structural model class can give rise to different estimators depending on the decision objective used in training.

\subsection{Additive Exponential Smoothing Class: Inference Results}
\label{sec:ETS_def_inference}

We consider the full class of Additive Exponential Smoothing Class  as well as the Simple Exponential Smoothing for clarity.

The formal definition of the time series process with all components:

\begin{align}
\mathsf{l}_t &= \alpha (y_t - \mathsf{s}_{t-m}) + (1 - \alpha) (\mathsf{l}_{t-1} + \mathsf{b}_{t-1}) \label{eq:hw_level} \\
\mathsf{b}_t &= \beta (\mathsf{l}_t - \mathsf{l}_{t-1}) + (1 - \beta) \mathsf{b}_{t-1} \label{eq:hw_trend} \\
\mathsf{s}_t &= \gamma (y_t - \mathsf{l}_{t-1} - \mathsf{b}_{t-1}) + (1 - \gamma) \mathsf{s}_{t-m} \label{eq:hw_seasonal}\\
y_{t+1} &= \mathsf{l}_t + \mathsf{b}_t + \mathsf{s}_{t+1-m} + \varepsilon_t 
\label{eq:hw} 
\end{align}

The states $\mathsf{l}_t$ represents the instantaneous level of the series (intercept), $\mathsf{b}_t$ an instantaneous linear trend and $\mathsf{s}_t$ the seasonal pattern that cycles with periodicity $m$. Parameters $\alpha, \beta, \gamma$ control the rate of change over time the states $\mathsf{l}_t, \mathsf{b}_t, \mathsf{s}_t$ respectively, belonging to the range $(0,1)$. Values closer to 0 imply no change over time (historical average) and closer to 1 total change over time.
The full model is known as Holt-Winters (HW), submodels such as no seasonality (a.k.a. Holt) or level-only (a.k.a Simple Exponential Smoothing) are obtained by setting their respective parameters and starting states to 0.

The experimental setup sets starting states $\mathsf{l}_0, \mathsf{b}_0$ and $\mathsf{s}_{-m+1:0}$ from the uniform $\mathrm{U}(-10,10)$ and parameters $\alpha, \beta, \gamma$ from $\mathrm{U}(0,1)$. Seasonal period $m$ is set to 4 and minimum length of the series is set to twice the seasonal period + number of parameters, (8+3=11) so that competitor estimators have enough degrees of freedom to fit. Maximum length of the series is set to 64, with lengths sampled uniformly. Innovations $\varepsilon_t$ come from a standard i.i.d. normal $\mathcal{N}(0,1)$.

We start with a smaller version of the experiment for the subclass of Simple Exponential Smoothing, a special case of Equation~\ref{eq:hw} setting $\beta, \gamma, \mathsf{b}_0, \mathsf{s}_{-m+1:0}=0$, then $\mathsf{l}_0$ to the standard normal, and fixed length=16. This simplification is done to isolate the effect of the estimators as much as possible, as well as to compare to the more computationally expensive Bayesian estimator.

Three neural estimators are trained, \textit{Neural-Unif} trained to minimize expected squared error under the uniform sampling of $\alpha$ (Section ~\ref{sec:training_algorithm});
\textit{Neural-Debias} trained with weights 0.05 for expected error and 0.95 for the squared bias penalty (Section~\ref{sec:frequentist_properties});  \textit{Neural-Minimax}, trained
to approximate a minimax risk estimator (Section~\ref{subsec:minimaxtrain}). The reference estimators are \textit{SES-MLE} maximum likelihood  implemented in python's package \texttt{statsmodels} and \textit{Bayes-Unif}, a Bayesian estimator of the posterior mean of SES implemented in \texttt{bayesforecast} in R, with the same priors and parameters as in the experiment.

\begin{figure}[htbp]
    \centering
    \begin{minipage}[t]{0.48\textwidth}
        \centering
        \includegraphics[width=\textwidth]{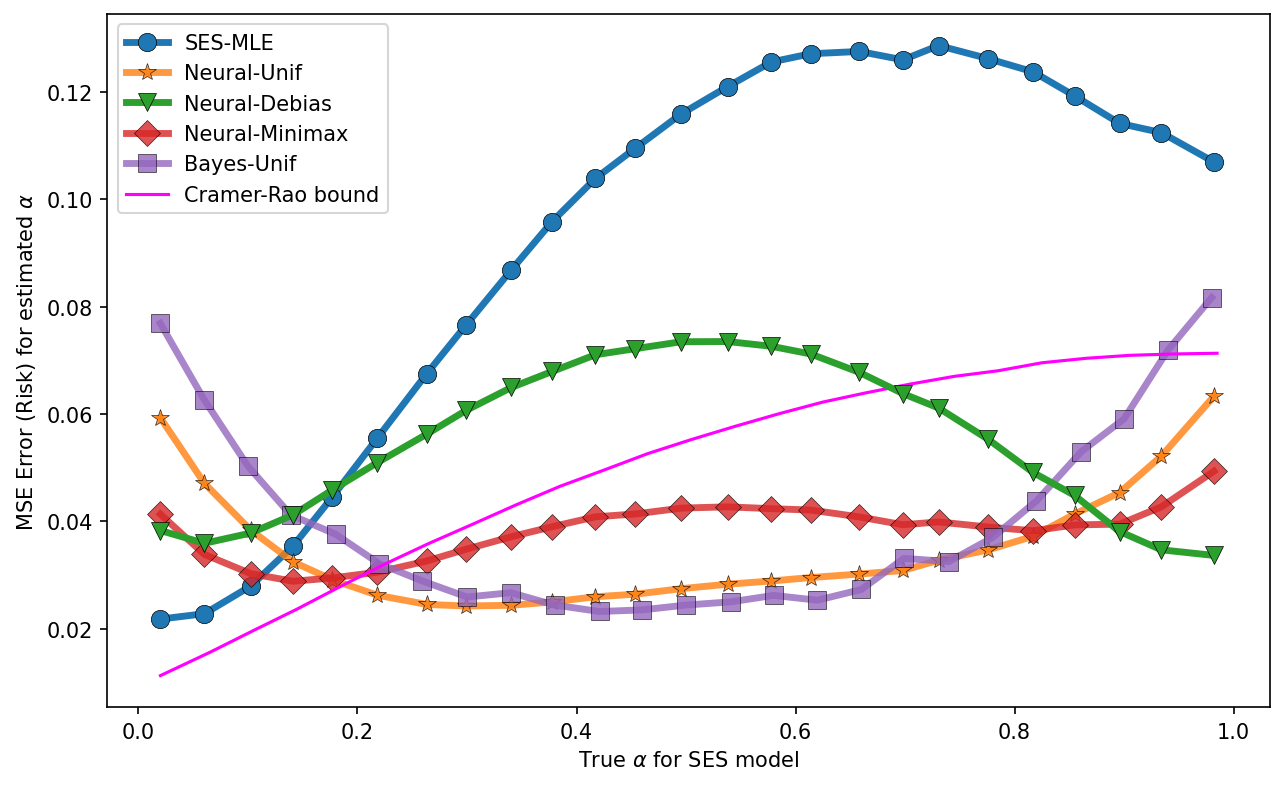}
    \end{minipage}
    \hfill
    \begin{minipage}[t]{0.48\textwidth}
        \centering
        \includegraphics[width=\textwidth]{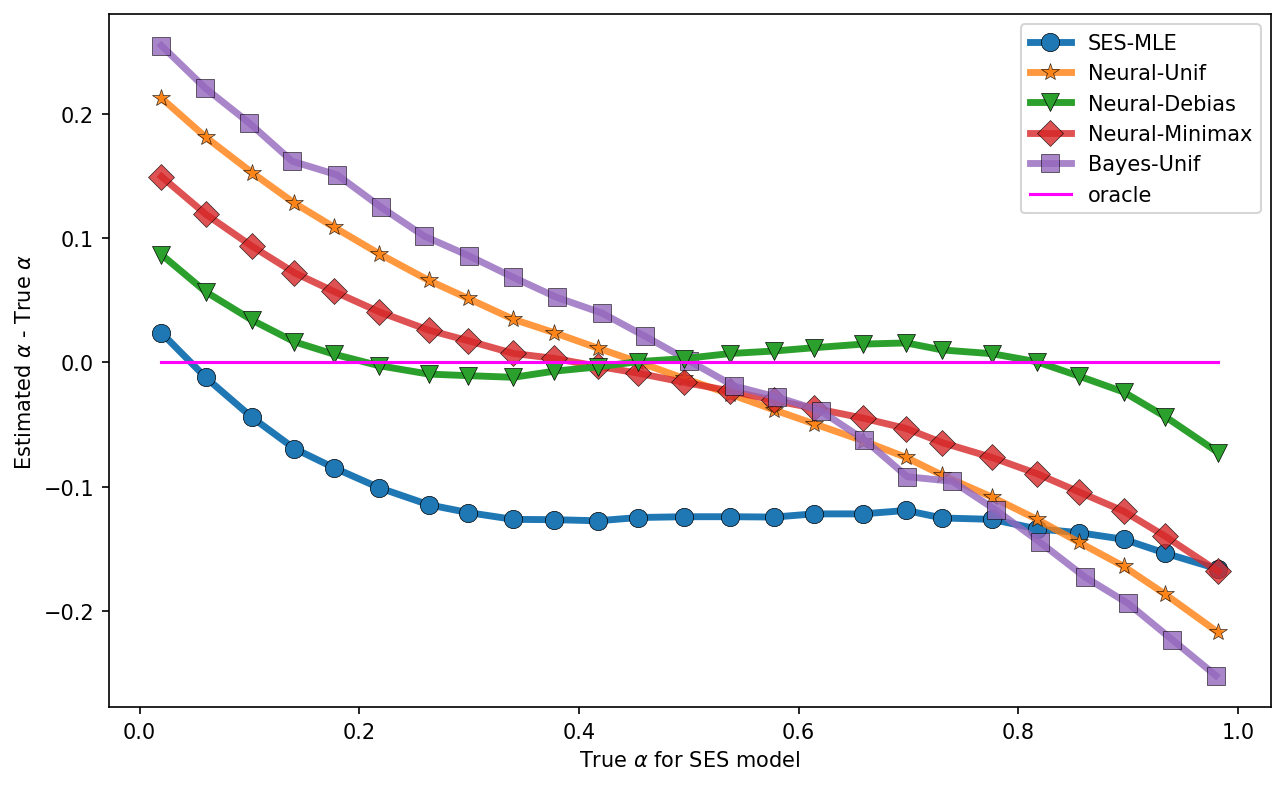}
    \end{minipage}
    \caption{Comparison of risk (left) and bias (right) for estimators of $\alpha$ in Simple Exponential Smoothing. 
    Neural-Unif has better overall performance than Maximum Likelihood and a Bayesian procedure. Neural-Debias pays a price in risk for reduced bias. Neural-Minimax has lower maximum risk than all other estimators. Maximum Likelihood has strong negative bias (towards 0). Other estimators bias toward 0.5, while Neural-Debias strongly reduces bias.}
    \label{fig:ses_alpha_side_by_side}
\end{figure}

Left panel in Figure~\ref{fig:ses_alpha_side_by_side} shows MSE per $\alpha$ for each of the estimators, on 1 Million time series, except the Bayes estimator that is run on 70000. The Cramer-Rao lower bound is shown in magenta for reference: the lower theoretical limit for error for unbiased estimators. Maximum Likelihood (SES-MLE) has good performance for values of $\alpha$ close to 0 (when the process resembles i.i.d. data) but has worse results outside of this range. Both Neural-Unif and Bayes-Unif are comparable, they aim to minimize expected error (Bayes Risk) over the uniform distribution of $\alpha$ (i.e. a Uniform prior) and we should expect to have the same performance. They show similar U-shaped risk curves, but the neural estimator is overall superior, everywhere but on the central values of $\alpha$ near 0.5. We attribute this difference to limitations of \texttt{bayesforecast} Markov Chain Monte Carlo under the uniform prior (such problems as the extremes of the range of values). The Neural-Debias estimator has much better performance at the extremes of $\alpha$ but trades off with a worse performance overall compared to Neural-Unif. The Neural-Minimax estimator succeeds in reducing the maximum error of any of the estimators, and the risk curve has become flatter, but it has three peaks at the extremes and center of the range of $\alpha$.  Importantly, no single estimator is dominated, and no estimator is universally better than the rest. Even SES-MLE outperforms the others in a limited range of $\alpha$. Numeric comparison is
shown in the Appendix Table~\ref{tab:ses_inference}.

Right panel in Figure~\ref{fig:ses_alpha_side_by_side} shows the results for estimation bias, centered for better visual appreciation (closer to 0 means better, less bias).
Maximum Likelihood shows a strong negative bias, consistently estimating lower $\alpha$ than the population value.
Neural-Unif and Bayes-Unif show bias towards $\alpha=0.5$: population $\alpha < 0.5$ are overestimated, $\alpha > 0.5$ are underestimated.
The bias of Bayes-Unif is stronger than the Neural-Unif, coherent with a better MSE of Bayes-Unif around $\alpha=0.5$ indicating stronger bias.
Neural-Minimax estimator has substantially reduced the bias compared to Neural-Unif.
The Neural-Debias has vastly reduced the overall bias of all the estimators though it is not completely unbiased.
Neural-Debias has lower squared bias than SES-MLE for all values of $\alpha$ above 0.1, and lower squared bias than the other estimators overall.
The MSE-bias loss tradeoff for this Neural-Debias estimator was set during training to $\lambda=0.05$ ($\text{bias}^2$ loss receives 19 times the weight of MSE). Better bias can be achieved with more penalty, but it cannot be completely reduced to 0 because $\alpha$ lies in a bounded range.

The inference results for the full experiments of the Holt-Winters model $\alpha, \beta, \gamma$ is summarized in Table~\ref{tab:hw_inference}.
The Neural-Debias estimator has successfully reduced the bias of Neural-Unif and shows a better
tradeoff than Maximum Likelihood. 
In conclusion, neural estimators achieved their objectives of optimizing for risk, bias reduction and lower worst-case errors. 

\begin{table}[htbp] 
\centering 
\caption{Comparison of Parameter Estimation Errors (MSE and Bias Squared) for Holt-Winters Exponential Smoothing. The neural estimators show better bias and squared error than Maximum Likelihood (HW-MLE)}
\label{tab:hw_inference}
\begin{tabular}{llrrr} 
\toprule
\multirow{2}{*}{Parameter} & \multirow{2}{*}{Error Type} & \multicolumn{3}{c}{Model} \\ 
\cmidrule(lr){3-5} 
                   &                             & HW-MLE & Neural-Unif & Neural-Debias \\
\midrule
\multirow{2}{*}{Alpha ($\alpha$)} & MSE       & 0.0602 & 0.0204 & 0.0606 \\
                   & $\text{bias}^2$    & 0.0113 & 0.0070 & 0.0003 \\
\midrule 
\multirow{2}{*}{Beta ($\beta$)}   & MSE       & 0.1620 & 0.0428 & 0.2498 \\
                   & $\text{bias}^2$    & 0.1180 & 0.0270 & 0.0045 \\
\midrule
\multirow{2}{*}{Gamma ($\gamma$)} & MSE       & 0.1343 & 0.0262 & 0.0621 \\
                   &  $\text{bias}^2$    & 0.0798 & 0.0091 & 0.0013 \\
\bottomrule
\end{tabular}
\end{table}

\subsection{Forecasting Scaling Laws in Exponential Smoothing}
\label{sec:ETS_scaling}
We now move into forecast performance, training a neural estimator per forecast horizon, for horizons $h=1, \ldots, 8$ ($H=8$). The neural estimators optimize for forecast risk (Mean Squared Error). We consider Simple Exponential Smoothing and Holt-Winters separately to gauge the effect of the increase in model complexity.
The neural estimators are set to have scale and location equivariance for the forecasts via min-max normalization,
to reduce the informativeness in the sampling. The reference estimators are Maximum Likelihood, SES-MLE and HW-MLE (in \texttt{statsmodels}).

\begin{figure}[H] 
    \centering

    \begin{minipage}[t]{0.48\textwidth}
        \centering
        \includegraphics[width=0.9\linewidth]{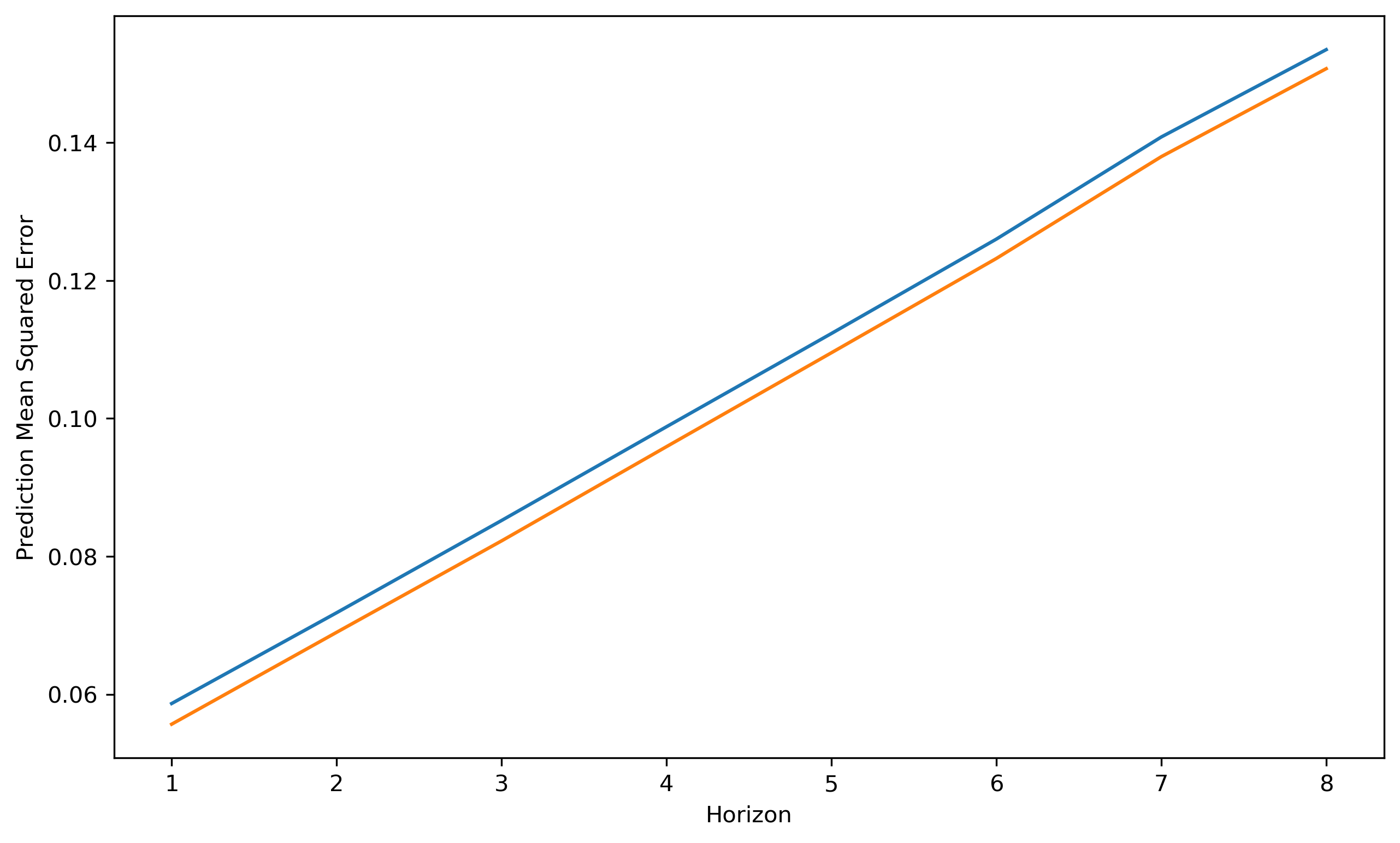}
    \end{minipage}
    \hfill
    \begin{minipage}[t]{0.48\textwidth}
        \centering
        \includegraphics[width=0.9\linewidth]{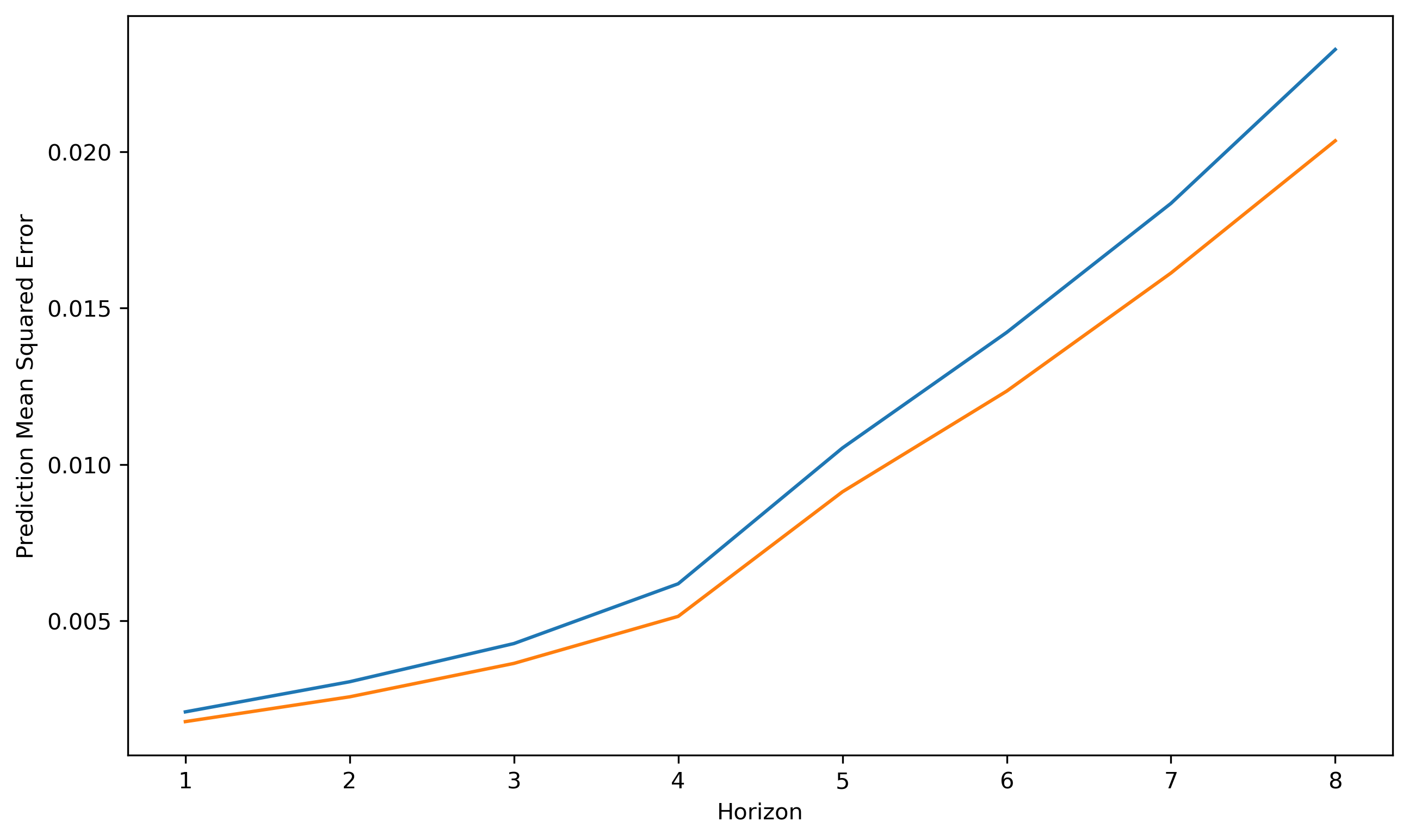}
    \end{minipage}

    \caption{Squared forecast error of neural estimators (orange) versus maximum likelihood (blue) as a function of the forecasting horizon. 
    The left panel corresponds to Simple Exponential Smoothing, while the right panel shows Holt--Winters.}
    \label{fig:ETS_forec_errors}
\end{figure}

Figure~\ref{fig:ETS_forec_errors} shows forecast MSE per horizon, for Simple Exponential Smoothing (left) and Holt-Winters (right), across 1M samples following the sampling mechanism in Section~\ref{sec:ETS_def_inference}. The neural estimator outperforms Maximum likelihood, and the evolution of performance with horizon is consistent with the theoretical variance of each process, linear with horizon for SES and quadratic for HW.

We consider two `scaling laws':
\begin{enumerate}
\item \textbf{Asymptotic performance based on series length}. Once the neural network estimator is trained, how it compares to MLE as the length of the series increases (sample size).
\item \textbf{Pre-training Dataset size scaling}. How many simulated time series are needed to train a neural estimator that outperforms Maximum Likelihood at different lengths for the series.
\end{enumerate}

\begin{figure}[H] 
    \centering

    \begin{minipage}[t]{0.48\textwidth}
        \centering
        \includegraphics[width=0.9\linewidth]{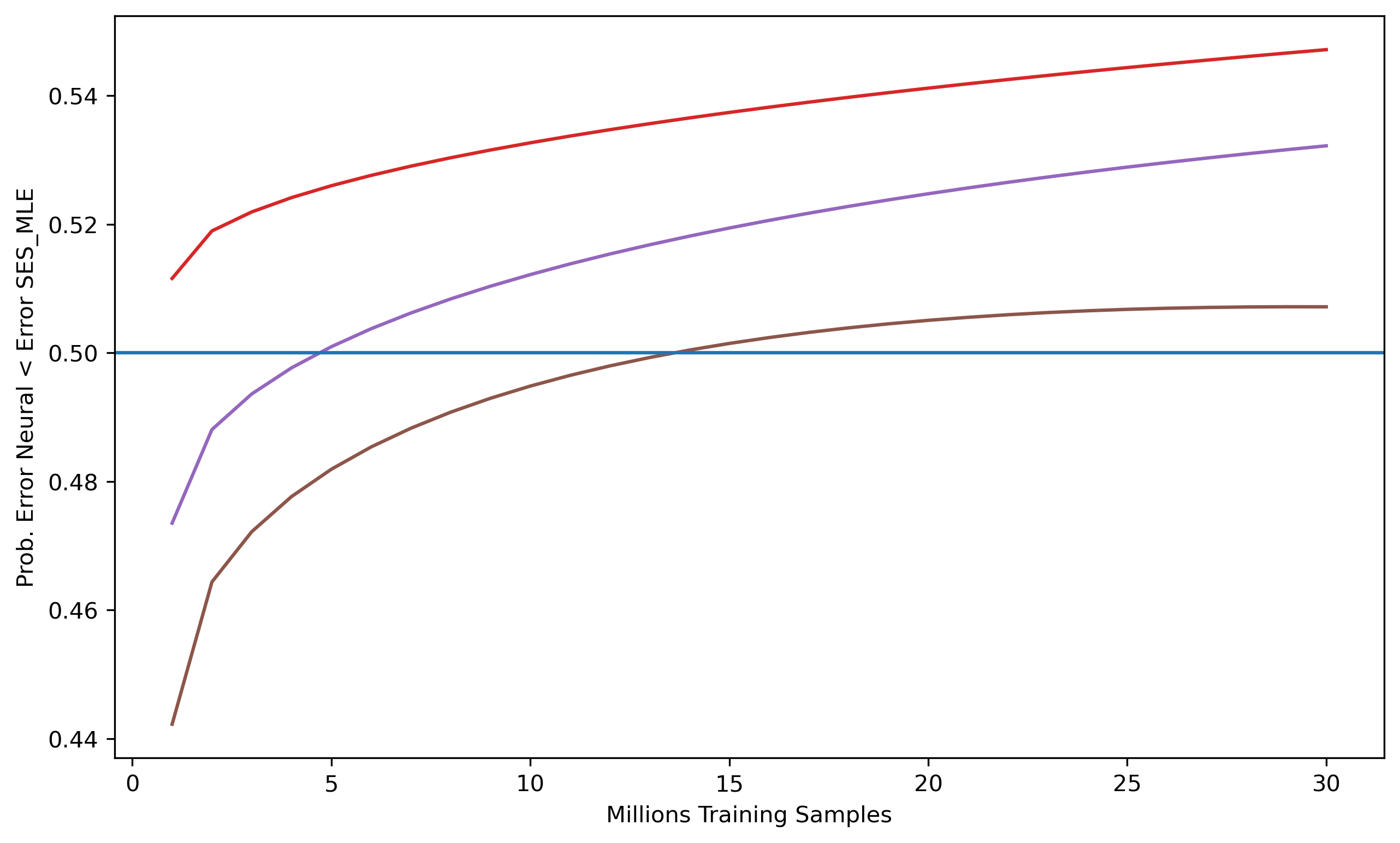}
    \end{minipage}
    \hfill
    \begin{minipage}[t]{0.48\textwidth}
        \centering
        \includegraphics[width=0.9\linewidth]{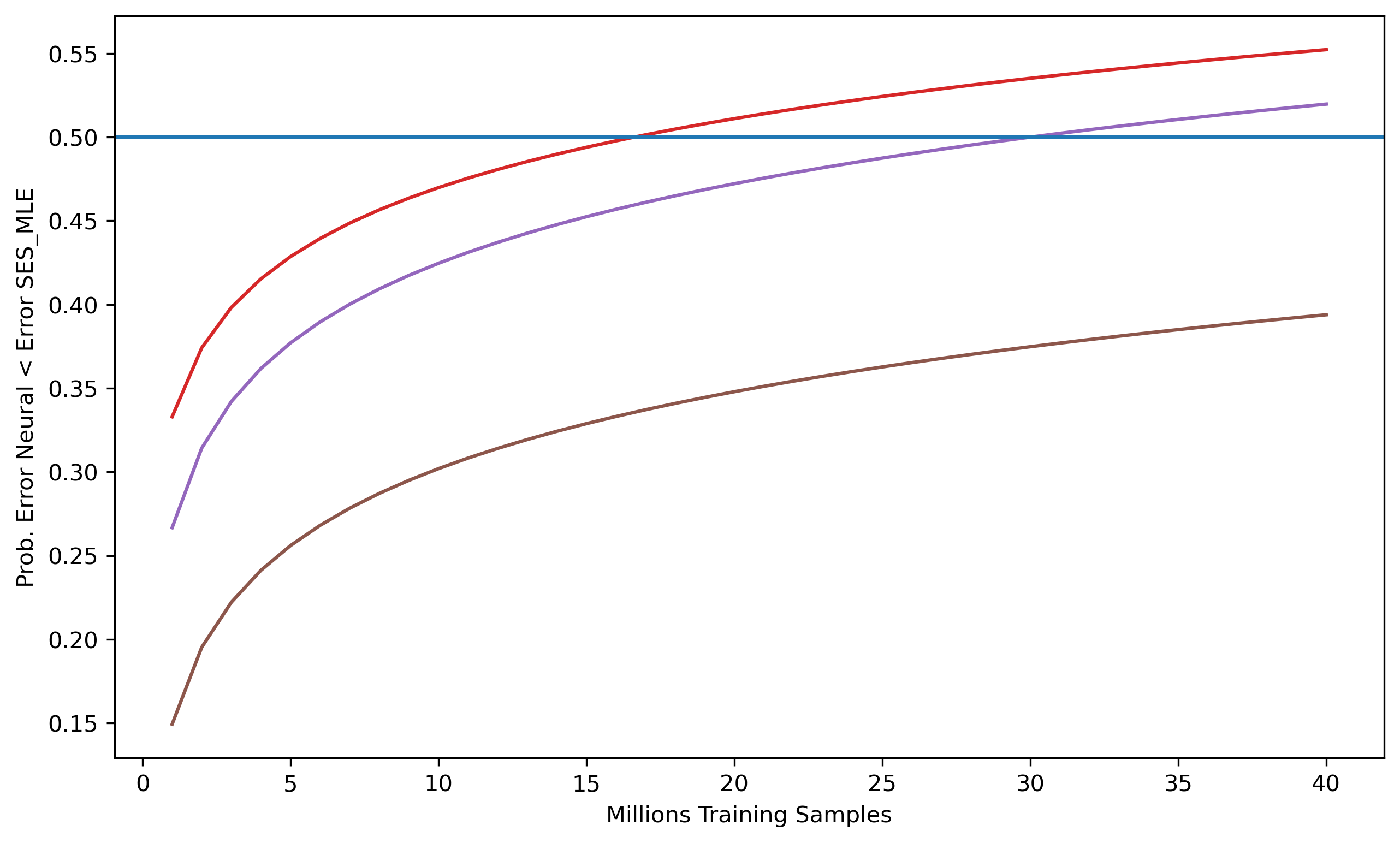}
    \end{minipage}

    \caption{Scaling laws for the forecast error performance of the Neural-Unif estimator relative to maximum likelihood. 
    The horizontal axis shows the training set size used for the neural estimator (simulated time series), while the vertical axis reports the percentage of wins, defined as the proportion of test cases in which the neural estimator attains lower forecast error than MLE. 
    Results for Simple Exponential Smoothing are shown on the left, and Holt--Winters on the right. 
    Each panel reports three series lengths (16, 32, and 64). Neural estimators can outperform MLE with sufficient training data, but require increasingly larger datasets as the series length grows and the underlying process becomes more complex (one parameter versus three parameters).}
    \label{fig:ETS_scaling_tourn}
\end{figure}

Figure~\ref{fig:ETS_scaling_tourn} shows how many training examples are required for
the neural estimator until it outperforms Maximum Likelihood. The series length are separated into 16, 32 and 64 to illustrate asymptotic properties.
We use the Percentage Wins metric, the proportion of series for which the neural estimator outperforms the reference.
This metric is more efficient when measuring small differences in MSE and more robust, accounts for numerical optimizer issues in the MLE estimators.
Experiment size is 20K time series for each length. Left panel in Figure~\ref{fig:ETS_scaling_tourn} show the scaling for SES and right panel for HW.
The horizontal line in each plot is set to 0.5, the proportion required to achieve superior performance.
Lines have been smoothed so comparisons are intended to be qualitative.
It can be seen on both Figures the increase in performance with more training and that shorter series require less training to outperform Maximum Likelihood. 
Comparing SES vs HW, it can be seen that SES is a simpler problem than HW for the neural networks, they require less training to achieve superior performance. We attribute this effect to the increase in complexity of the process (more parameters) as well as the increased Signal to Noise ratio in HW, an artifact of the simulation mechanism.

It can be inferred from the plots that for very long series, Maximum Likelihood would be difficult to beat, confirming that at the asymptotic regime, the classic estimators are enough.
In this regime, differences between estimators become a numeric issue.
In the Appendix, Figure~\ref{fig:ses_length_mse} we measure forecast MSE, it can be confirmed that both neural and MLE have `converged' to the same square error performance long before the maximum length series in the experiment.

\subsection{Real Data: Forecasting Results of Neural Exponential Smoothing}
\label{sec:ETS_empiric}
We compare a neural estimator trained by the simulation scheme in Section~\ref{sec:ETS_def_inference} and we
test it on the M1 competition real dataset \cite{makridakis1982accuracy} with 1001 time series for Yearly, Quarterly and Monthly frequencies.
We adapt the simulation scheme for the M1: we add seasonality $m=12$ to cover Monthly data, sampling Yearly, Quarterly and Yearly with equal probability.
The estimator Neural-ETS, is trained to optimize for direct forecast MSE on horizons $h=1, \ldots, 18$ ( Monthly $H=18$, Quarterly $H=8$, Yearly $H=6$).
We choose the M1 because in their seminal paper \cite{hyndman2002state}, Hyndman et al. compare between a range of estimators for the same ETS class,
so it is a good way of isolating the effect of estimation and highlight the benefits of the Simulacrum approach.
Specifically, they implement 5 estimators: the Maximum Likelihood estimator (MLE), Least Squares estimator (MSE),
the estimator that minimizes in-sample Mean Absolute Percentage Errors (MAPE), the estimator that minimizes the residual variance (Sigma) and the estimator that minimizes the average of 1 to 3 step-ahead in-sample forecast errors (AMSE).

Results in Table~\ref{tab:estimators_mape_m1} show that the neural estimator achieves the best accuracy.
Ranking the estimators, a pattern emerges: top estimators are Neural, AMSE and MSE all optimize for the criteria of forecast error across horizons (unlike MLE).
AMSE gets better performance than MSE because its estimation objective is closer to the true objective in the M1 competition of average error across horizons 1 to 18.
The neural estimator can target this objective easily, even considering a set of models and does not need to rely on model selection that does not optimize directly for forecast error.
A limitation in the comparison is that Neural-ETS restricts to the additive class and two observations were removed due to the MAPE metric being unstable (division by 0).

\begin{table}[htbp] 
\centering
\caption{Comparison of Estimation Methods by MAPE for several ETS estimators in the M1 real data benchmark dataset. Neural-ETS Estimator considering all additive subclasses is the top performing method, followed by the estimator that optimizes for multi-horizon forecast error (AMSE). Table is an extension from Table 3 in \cite{hyndman2002state}.}
\label{tab:estimators_mape_m1}
\begin{tabular}{@{}lr@{}} 
\toprule
Estimation Method & \multicolumn{1}{c}{MAPE} \\ 
\midrule
AMSE   & 17.63 \\
MSE    & 17.73 \\
Sigma  & 18.49 \\
MLE    & 18.55 \\
MAPE   & 19.08 \\
\midrule
Neural & 17.33 \\
\bottomrule
\end{tabular}
\end{table}

Extending the comparison beyond ETS estimators to the complete set of models that participated in the M1 competition and the recent TSFM Chronos-2 \cite{ansari2025chronos2}, we see that the neural estimator is very competitive. For the robust median APE metric used in the M1 paper, the neural estimator outperforms all other 21 models not only on average, but for most of the horizons, shown in Table~\ref{tab:median_ape_M1_all}. 
This is remarkable, many of the models that participated in the competition are variants of Exponential Smoothing, including more sophisticated alternatives,
while the neural estimator is just an alternative estimator of the classic additive class.
Results on average APE are also superior (though not for all horizons), seen in the Appendix Table~\ref{tab:avg_mape_all_data_no_avg_fh}.

\begin{table}[htbp] 
\centering
\caption{Median APE: all M1 Competition data (1001). The Neural-ETS estimator outperforms all other methods that participated in the M1 competition.}
\label{tab:median_ape_M1_all}
\resizebox{0.8\textwidth}{!}{
\begin{tabular}{@{}l rrrrrrrrrr@{}} 
\toprule
& \multicolumn{10}{c}{FORECASTING HORIZONS} \\
\cmidrule(lr){2-11} 
METHODS & \multicolumn{1}{c}{1} & \multicolumn{1}{c}{2} & \multicolumn{1}{c}{3} & \multicolumn{1}{c}{4} & \multicolumn{1}{c}{5} & \multicolumn{1}{c}{6} & \multicolumn{1}{c}{8} & \multicolumn{1}{c}{12} & \multicolumn{1}{c}{15} & \multicolumn{1}{c}{18} \\
\midrule
NAIVE 1      & 6.0 & 8.8 & 9.1 & 10.8 & 12.0 & 13.3 & 12.2 & 10.4 & 13.9 & 15.7 \\
MOV.AVERAG   & 6.3 & 8.9 & 9.0 & 10.7 & 11.7 & 13.1 & 11.4 & 11.8 & 13.6 & 15.4 \\
SINGLE EXP   & 6.0 & 8.2 & 8.6 & 10.5 & 11.2 & 12.5 & 11.1 & 11.3 & 13.4 & 14.9 \\
ARR EXP      & 6.7 & 9.7 & 8.5 & 10.2 & 11.2 & 12.7 & 11.2 & 12.2 & 13.4 & 14.7 \\
HOLT EXP     & 6.1 & 7.2 & 7.1 & 8.9  & 9.5  & 11.4 & 11.5 & 11.3 & 14.2 & 17.2 \\
BROWN EXP    & 5.7 & 6.9 & 7.6 & 8.6  & 10.0 & 11.8 & 12.0 & 14.6 & 15.5 & 19.1 \\
QUAD. EXP    & 6.3 & 7.8 & 8.9 & 9.8  & 12.0 & 14.3 & 16.2 & 19.7 & 25.8 & 29.9 \\
REGRESSION   & 9.0 & 9.9 & 9.6 & 10.8 & 11.0 & 12.2 & 11.7 & 12.6 & 13.1 & 14.1 \\
NAIVE2       & 4.8 & 6.6 & 6.9 & 8.8  & 9.9  & 10.8 & 10.4 & 10.4 & 12.7 & 12.6 \\
D MOV.AVRG   & 5.8 & 7.9 & 9.5 & 11.6 & 12.3 & 13.8 & 11.5 & 11.5 & 14.1 & 16.4 \\
D SING EXP   & 4.7 & 6.2 & 6.8 & 8.2  & 9.0  & 10.5 & 10.0 & 10.3 & 12.1 & 12.5 \\
D ARR EXP    & 5.3 & 7.3 & 7.1 & 8.6  & 9.6  & 10.9 & 10.0 & 10.6 & 12.6 & 12.9 \\
D MULT EXP   & 4.5 & 5.3 & 5.6 & 7.3  & 8.1  & 9.2  & 9.8  & 9.9  & 12.2 & 13.8 \\
D BROWNEXP   & 4.4 & 5.1 & 6.2 & 7.3  & 8.5  & 9.6  & 10.0 & 11.1 & 12.7 & 15.9 \\
D QUAD.EXP   & 4.5 & 5.7 & 6.9 & 8.3  & 10.0 & 11.7 & 13.9 & 15.6 & 20.4 & 24.8 \\
D REGRESS    & 7.8 & 7.9 & 9.1 & 9.1  & 9.8  & 11.2 & 10.6 & 10.9 & 12.6 & 11.9 \\
WINTERS      & 4.4 & 5.4 & 5.5 & 6.6  & 7.9  & 9.6  & 9.7  & 10.1 & 11.0 & 12.2 \\
AUTOM. AEP   & 4.7 & 5.8 & 6.6 & 7.2  & 8.5  & 9.7  & 9.6  & 10.5 & 12.4 & 14.1 \\
BAYESIAN F   & 5.1 & 5.5 & 6.4 & 7.8  & 8.5  & 10.0 & 10.2 & 10.1 & 12.5 & 13.1 \\
COMBINING A  & 4.2 & 5.4 & 5.9 & 7.1  & 8.0  & 9.7  & 9.3  & 9.5  & 11.8 & 12.6 \\
COMBINING R  & 4.6 & 6.0 & 6.6 & 7.7  & 8.8  & 10.1 & 10.1 & 10.4 & 12.3 & 12.8 \\
\midrule
Chronos-2    & 5.2 & 6.5 & 6.6 & 8.5  & 9.7  & 10.9 & 9.6 & 10.1 & 10.9 & 12.2 \\
\midrule
Neural-ETS  & 4.3 & 5.1 & 5.6 & 6.9  & 7.9  & 8.9 & 9.0 & 9.1 & 10.8 & 11.9 \\
\bottomrule
\end{tabular}%
}
\end{table}

\subsection{Real Data: Performance of Neural-ETS Minimax Estimator}
\label{sec:ETS_minimax}

Assessing performance of a minimax estimator in real data is more difficult than in simulation, because the true parameters are unobserved and forecast errors contain substantial irreducible noise.
However, if the dataset is sufficiently heterogeneous and includes challenging series, a minimax estimator's properties could be appreciated in the \emph{distribution} of forecast errors: relative to an estimator optimized for average risk, it should produce fewer extremely large errors, at the cost of somewhat worse performance on easier cases.

\begin{figure}[H] 
    \centering

    \begin{minipage}[t]{0.49\textwidth}
        \centering
        \includegraphics[width=1.05\textwidth]{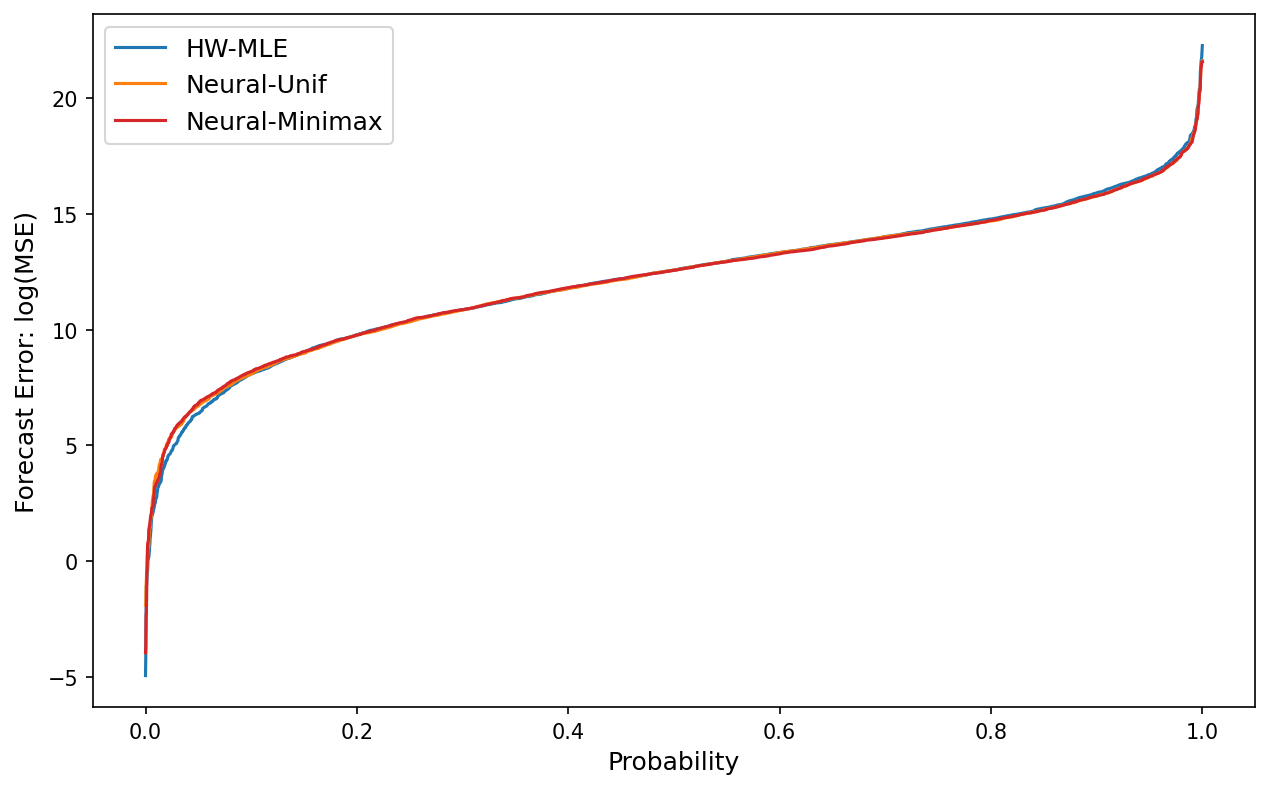}
    \end{minipage}
    \hfill
    \begin{minipage}[t]{0.49\textwidth}
        \centering
        \includegraphics[width=1.05\textwidth]{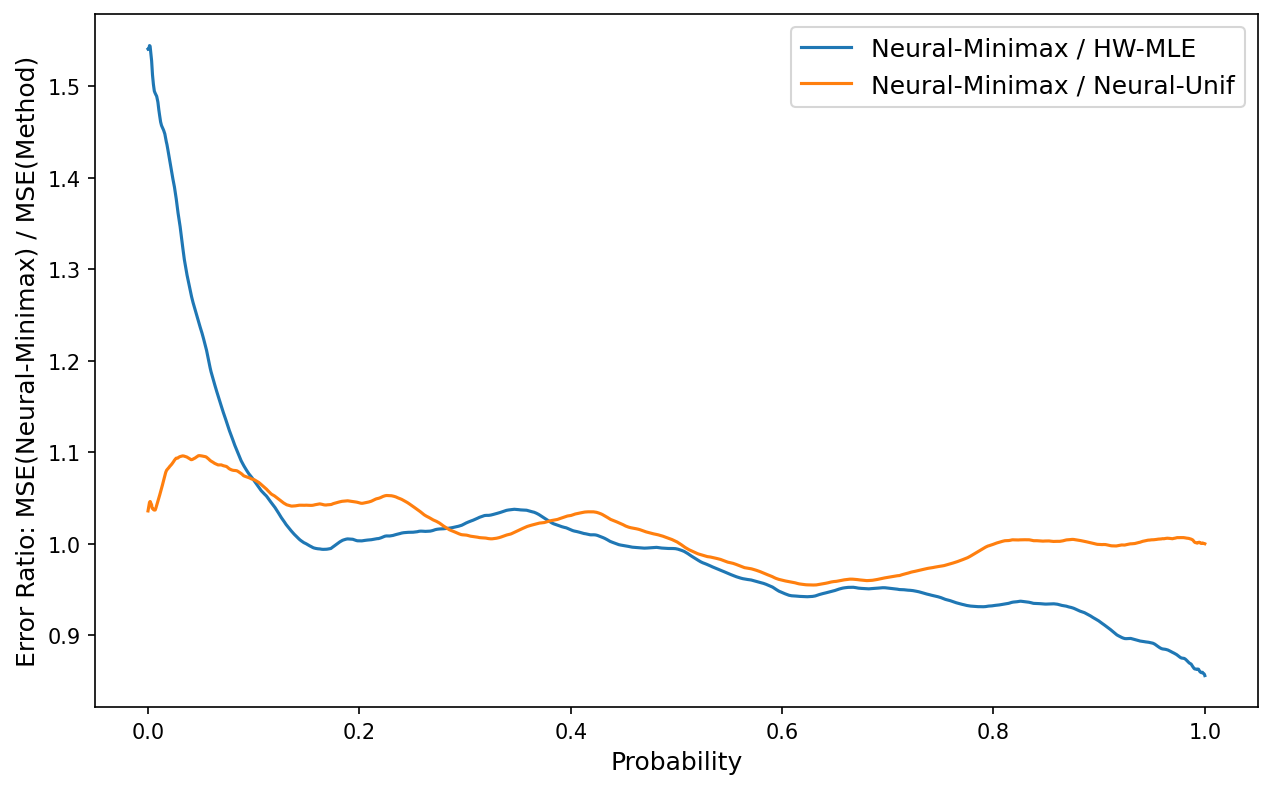}
    \end{minipage}

    \caption{Forecast-error distribution for HW-MLE, Neural-Unif, and Neural-Minimax on the M3 Yearly dataset.
    The left panel shows the empirical quantile functions (EQFs) of log forecast MSE.
    The right panel shows smoothed ratios of the EQF of Neural-Minimax relative to HW-MLE and Neural-Unif.
    Values below 1 indicate smaller errors for the minimax estimator.
    Neural-Minimax is slightly worse at low quantiles but improves in the upper tail, consistent with its goal of reducing extreme forecast errors.}
    \label{fig:quantile_error_dist}
\end{figure}

We examine this trade-off using the 675 yearly series in the M3 dataset.
Yearly series provide a useful test bed because they are volatile enough to generate extreme forecast failures.
The left panel of Figure~\ref{fig:quantile_error_dist} plots the empirical quantile functions (EQFs) of forecast MSE for HW-MLE, Neural-Unif, and Neural-Minimax.
For a given probability level $p$, the EQF $\hat{Q}(p)$ reports the error threshold below which a proportion $p$ of the observations fall; equivalently, larger values of $p$ correspond to increasingly extreme errors.

The left panel suggests that Neural-Minimax and Neural-Unif have similar overall profiles relative to HW-MLE, as expected from the risk profiles studied in Section~\ref{sec:ETS_def_inference}.
To make the tail comparison clearer, the right panel of Figure~\ref{fig:quantile_error_dist} reports the ratios
\[
\frac{\hat{Q}_{\mathrm{MSE}(\mathrm{Neural\text{-}Minimax})}(p)}
     {\hat{Q}_{\mathrm{MSE}(\mathrm{HW\text{-}MLE})}(p)}
\quad \text{and} \quad
\frac{\hat{Q}_{\mathrm{MSE}(\mathrm{Neural\text{-}Minimax})}(p)}
     {\hat{Q}_{\mathrm{MSE}(\mathrm{Neural\text{-}Unif})}(p)}.
\]
Values above 1 indicate worse performance for Neural-Minimax, while values below 1 indicate an advantage. The plot has been smoothed for clarity, the results should be taken qualitatively.

The pattern is consistent with the minimax objective.
At lower quantiles, Neural-Minimax has slightly larger errors than HW-MLE, indicating some sacrifice in best-case performance.
However, from approximately $p \geq 0.5$ onward, its relative performance improves, and in the upper tail it produces smaller errors.
At the extreme, the largest forecast error of Neural-Minimax is about 85\% of the largest error of HW-MLE.
The comparison with Neural-Unif is qualitatively similar but less pronounced.
Overall, these results suggest that Neural-Minimax behaves as intended: it trades off some performance on easier cases in exchange for better protection against extreme forecast failures.

\section{Application II: Solving the Bias and Calibration Problem in AR(p) Models}
\label{sec:ARp}

The autoregressive process of order $p$, denoted AR($p$), is a fundamental model in time series analysis. Beyond forecasting applications, AR structures are critical in regression modeling to characterize correlated errors (e.g., \cite{lichstein2002spatial}). In such settings, neglecting autocorrelation compromises inference, yielding incorrect parameter estimates, biased standard errors, and invalid hypothesis tests \cite{cheang2000bias}.

Formally, an AR($p$) process generates observations recursively as a linear combination of past values, with coefficients $\phi_1, \dots,\phi_p$, plus an intercept $c$ and an innovation term $\varepsilon_t$ (typically i.i.d. Gaussian):
\begin{align}
y_{t+1} &= c + \phi_1 y_t + \cdots + \phi_{p}y_{t-p +1} + \varepsilon_t
\label{eq:arp}
\end{align}

With sufficient order $p$, these models can approximate highly complex dynamics, effectively capturing persistence and dependence. However, despite their linearity, statistical estimation is non-trivial due to the temporal dependency of observations. Standard approaches face inherent limitations: Ordinary Least Squares (OLS) is inefficient, while the ``gold standard'' Exact Maximum Likelihood (MLE) suffers from well-documented finite-sample bias.

This estimation bias severely limits applicability in scientific domains where rigorous unbiasedness is critical. Furthermore, point estimation bias is compounded by the challenge of quantifying \textbf{parameter uncertainty}. Standard estimators fail to achieve \textbf{uniform calibration}: validity across the entire parameter space. Instead, they often yield confidence intervals that are valid on average but systematically invalid for specific parameter configurations. While considerable research has focused on bias correction (dating back to 1954 \cite{marriott1954bias}, with recent advances for low-order processes \cite{sorbye2022finite} and machine learning-assisted expansions \cite{muller2024bias}), achieving uniform calibration remains more challenging.

This section has two goals:

\begin{itemize}
    \item compare point-estimation risk and bias for neural and classical estimators in AR(5), including maximum likelihood, OLS, Yule--Walker, and Burg estimators (Section~\ref{sec:arpriskbias});
    \item evaluate distributional forecasting and calibration, including marginal calibration, uniform calibration across the parameter space, and simultaneous forecast-path coverage (Section~\ref{sec:arpcalibr}).
\end{itemize}

\subsection{Estimation error and bias of Neural Estimators for the AR(p) class}
\label{sec:arpriskbias}

We compare performance of two neural estimators: optimizing for minimum expected squared error (Neural-Unif) and reducing the bias (Neural-Debias).
Our reference estimators are Burg (AR-BURG), Yule-Walker (AR-YW), Conditional Least Squares (AR-OLS) and
Exact Maximum Likelihood (AR-MLE), implemented in \texttt{stats} core package in \texttt{R}.

For the experimental setup to train and evaluate the neural estimators, we restrict the order of AR to $p=5$.
To sample from the parameter space of stationary AR processes we first sample $p$ Partial Autocorrelation coefficients from the $p$-dimensional uniform hypercube $\mathrm{U}(-1,1)^p$.
The sampled partial autocorrelations are then transformed into AR coefficients via the Durbin-Levinson, enforcing coefficients to the stationary subspace.
Innovation process is set to $\mathrm{N}(0,1)$ innovations, constant $c$ is set to 0.
Series lengths vary uniformly from length 12 ( 2 times $p$ + 2, so estimation is feasible) to 64.
The reference AR models are correctly specified according to the simulation mechanism, fixing the order to 5 and disabling the estimation of the additive constant.
The neural estimators are made location and scale invariant so that the variance of the innovation process is not informative.

The comparison of the estimators averaging across the parameter space is shown in Table~\ref{tab:ar5_inference_errors} over a test set of 1M sample size.
Neural estimators are superior to all other estimators in terms of expected squared error (MSE) and bias. The result holds both when averaging across the sampling mechanism (the inversions of the PACF hypercube) but also for each single coefficient. Neural-Debias has succeeded in reducing the bias of Neural-Unif at a tradeoff in MSE. Among the reference estimators, Maximum Likelihood has the best performance, while Yule-Walker and Least Squares have relatively poor performance. This might be attributed to numeric issues that affect the squared error metric. AR-MLE was also prone to numeric issues with convergence of the optimization method used in the estimation, in about 10\% of the time series, when errors were raised, the fallback method was set to AR-BURG. 

\begin{table}[htbp]
\setlength{\tabcolsep}{4pt}
\centering
\caption{Inference errors MSE and $\text{bias}^2$ for estimation of coefficients in an AR(5) for neural and classic estimators, averaging across each of the parameters. The Neural-Unif estimator achieves best overall performance for all dimensions, the Neural-Debias effectively reduces the bias. The best classic estimator is Exact Maximum Likelihood (AR-MLE) followed by Burg (AR-BURG)}
\label{tab:ar5_inference_errors}
\begin{tabular}{llrrrrrr}
\toprule
\multirow{2}{*}{Coefficient} & \multirow{2}{*}{Error Type} & \multicolumn{6}{c}{Method} \\
\cmidrule(lr){3-8}
 & & AR-MLE & Neural-Unif & Neural-Debias & AR-OLS & AR-BURG & AR-YW \\ \midrule
\multirow{2}{*}{$\phi_1$} & MSE & 0.0526 & 0.0398 & 0.0440 & 0.1107 & 0.0524 & 0.1793 \\
 & $\text{bias}^2$ & 0.0048 & 0.0025 & 0.0005 & 0.0096 & 0.0043 & 0.0914 \\ \midrule
\multirow{2}{*}{$\phi_2$} & MSE & 0.0855 & 0.0603 & 0.0683 & 0.1616 & 0.0867 & 0.3260 \\
 & $\text{bias}^2$ & 0.0083 & 0.0048 & 0.0004 & 0.0152 & 0.0146 & 0.2146 \\ \midrule
\multirow{2}{*}{$\phi_3$} & MSE & 0.1426 & 0.0727 & 0.0802 & 0.2101 & 0.1041 & 0.4727 \\
 & $\text{bias}^2$ & 0.0336 & 0.0092 & 0.0001 & 0.0368 & 0.0287 & 0.3734 \\ \midrule
\multirow{2}{*}{$\phi_4$} & MSE & 0.1173 & 0.0761 & 0.0816 & 0.1756 & 0.1509 & 0.5385 \\
 & $\text{bias}^2$ & 0.0336 & 0.0108 & 0.0004 & 0.0341 & 0.0463 & 0.4519 \\ \midrule
\multirow{2}{*}{$\phi_5$} & MSE & 0.0548 & 0.0416 & 0.0454 & 0.1046 & 0.0536 & 0.1641 \\
 & $\text{bias}^2$ & 0.0141 & 0.0052 & 0.0007 & 0.0136 & 0.0134 & 0.1233 \\ \midrule
\bottomrule
\end{tabular}
\end{table}

\begin{figure}[htbp]
    \centering
    \begin{subfigure}[b]{0.48\textwidth}
        \centering
        \includegraphics[width=\textwidth]{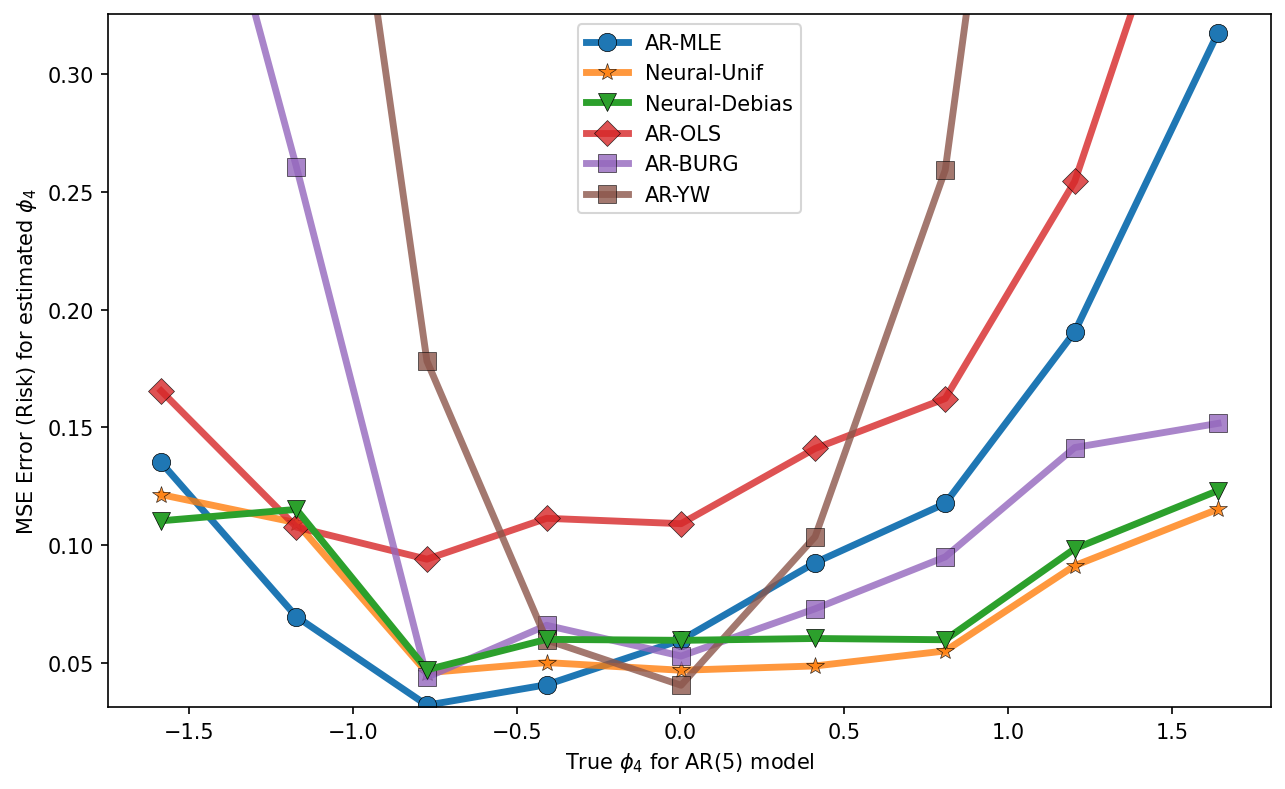}
        \caption{Estimation MSE vs $\phi_4$ in AR(5).}
        \label{fig:ar5_phi4_mse}
    \end{subfigure}
    \hfill 
    \begin{subfigure}[b]{0.48\textwidth}
        \centering
        \includegraphics[width=\textwidth]{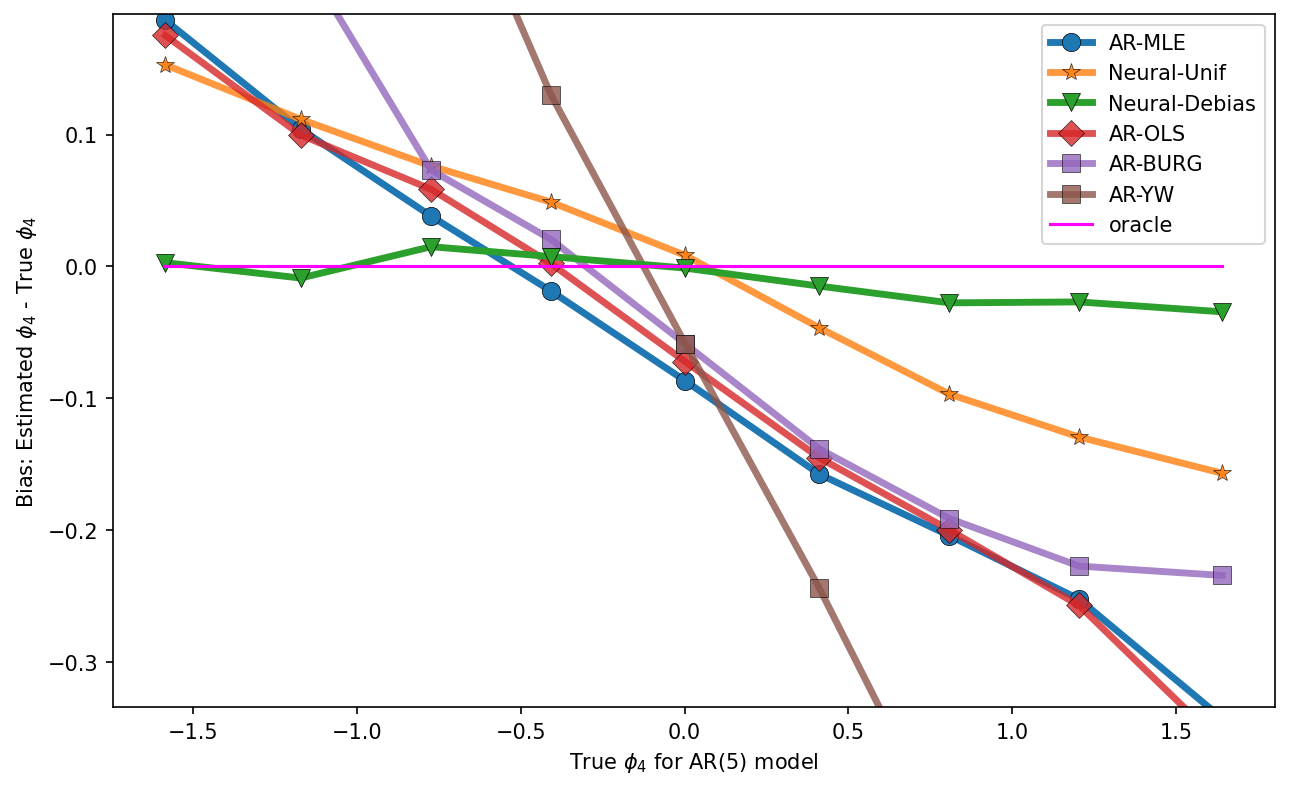}
        \caption{Estimation bias vs $\phi_4$ in AR(5).}
        \label{fig:ar5_phi4_bias}
    \end{subfigure}
    
    \caption{Detailed performance comparison for parameter $\phi_4$ in the AR(5) class. Left panel: Estimation MSE for each value of $\phi_4$ parameter in an AR(5) model. Classic and Neural estimators for the AR(5) are compared. Neural-Unif achieves best overall performance but Maximum Likelihood is better for $\phi_4 \leq -0.5$. The Neural-Debias estimator has overall worse MSE, trading off error for bias. Right panel: Estimation bias for each value of $\phi_4$ parameter in an AR(5) model, centered (closer to 0 is better). Classic and Neural estimators for the AR(5) are compared. The Neural-Debias estimator strongly reduces bias over all other alternatives, which exhibit bias towards 0 (or around 0).}
    \label{fig:ar5_phi4_combined}
\end{figure}

To showcase performance per value of a parameter, we focus on $\phi_4$, because smaller-order $\phi$ are more studied elsewhere.
Results for other coefficients are analogous.
MSE for the range of $\phi_4$ is depicted in Figure~\ref{fig:ar5_phi4_mse}.
We see that no estimator completely dominates all others, AR-MLE is superior to Neural-Unif for values of $\phi_4 \leq 0$,
so there is a range of the parameter space for which AR-MLE is better than the Neural-Unif, though Neural-Unif is clearly superior.
AR-MLE dominates AR-OLS. AR-YW, has very poor performance on average, but has the best performance in a neighbourhood around $\phi_4=0$.
The Neural-Debias estimator has good performance but it is virtually dominated by Neural-Unif, paying the price of reducing the bias.
The bias comparison is shown in Figure~\ref{fig:ar5_phi4_bias}.
The Neural-Debias estimator virtually dominates all other methods in terms of bias across the range of values.
All estimators show a bias towards 0 (AR-YW very strongly),
but AR-MLE has a smaller bias that Neural-Unif in the range of negative values of $\phi_4$ and larger bias for greater values.
We see that the relationship between estimators is complex, it cannot be summarized by simple rules.

In sum, the neural estimators have succeeded in reaching superior expected error and bias reduction for the main class of time series models in Time Series Analysis. Distributional forecast accuracy in the AR(p) is covered in the following Section~\ref{sec:arpcalibr}.

\subsection{Uniform Calibration and Distributional Accuracy of Neural Estimators for the AR(p) class}
\label{sec:arpcalibr}

Having addressed parameter estimation, we now turn to uncertainty quantification (which subsumes point estimation and is more challenging to solve).
The main result of this section is that our framework achieves strong distributional performance in the AR$(p)$ class, including accurate quantile forecasts, marginal calibration, simultaneous prediction paths, and, most importantly, \textbf{uniform calibration}.
Uniform calibration is particularly important in high-stakes settings, where risk must be controlled for each process rather than only on average across a population:

\begin{itemize}
    \item \textbf{Financial Regulation:} The Basel Committee's MAR32 framework requires backtesting calibration not just bank-wide, but at the granular trading-desk level \cite{BIS2023MAR32}.
    \item \textbf{Cloud Reliability:} Providers like Google set Service Level Objectives (SLOs) (e.g., 99.9\% uptime) that must hold for individual availability zones, not just the global average \cite{google2024SLO}.
    \item \textbf{Healthcare:} Clinical decision-making requires calibration valid for the specific patient profile, not just the average patient \cite{van2019calibration}. This parallels the distinction between `strong calibration' (uniform) and `mean calibration' (marginal).
\end{itemize}

Standard versions of the main forecasting paradigms do not generally target this property directly:

\begin{itemize}
    \item \textbf{Classical Frequentist:} Estimators like Maximum Likelihood or OLS are biased in finite samples. Consequently, prediction intervals are centered incorrectly, leading to a failure in uniform calibration.
    \item \textbf{Bayesian Methods:} Bayesian credible intervals guarantee \textit{average} calibration with respect to the prior distribution (Bayes Risk), not uniform calibration for every fixed parameter $\theta$.
    \item \textbf{Global \& Foundation Models:} Models like N-BEATS \cite{oreshkin2020nbeats} or Chronos \cite{ansari2024chronos} are trained via Empirical Risk Minimization over massive corpora. They optimize an aggregate objective, learning a ``shared'' uncertainty model that works best on average (\textit{Marginal Calibration}). Achieving uniform calibration in end-to-end training is difficult because it imposes constraints on individual series, effectively negating the benefits of shared data learning. Furthermore, individual series often lack sufficient data to train such complex models in isolation.
\end{itemize}

Our framework addresses this by simulating the parameter space directly, allowing us to penalize miscalibration at the individual process level. We compare three neural variants: \textbf{Neural-Unif} (standard Pinball loss), \textbf{Neural-Calibr} (calibration-aware loss), and \textbf{Neural-Simult} (simultaneous bands), against classical baselines. Note that \textbf{Neural-Unif} serves as a proxy for a standard Global/Foundation/PFN model trained on synthetic data.

\begin{figure}[htbp]
    \centering

    \begin{subfigure}[t]{0.49\textwidth}
        \centering
        \includegraphics[width=\linewidth]{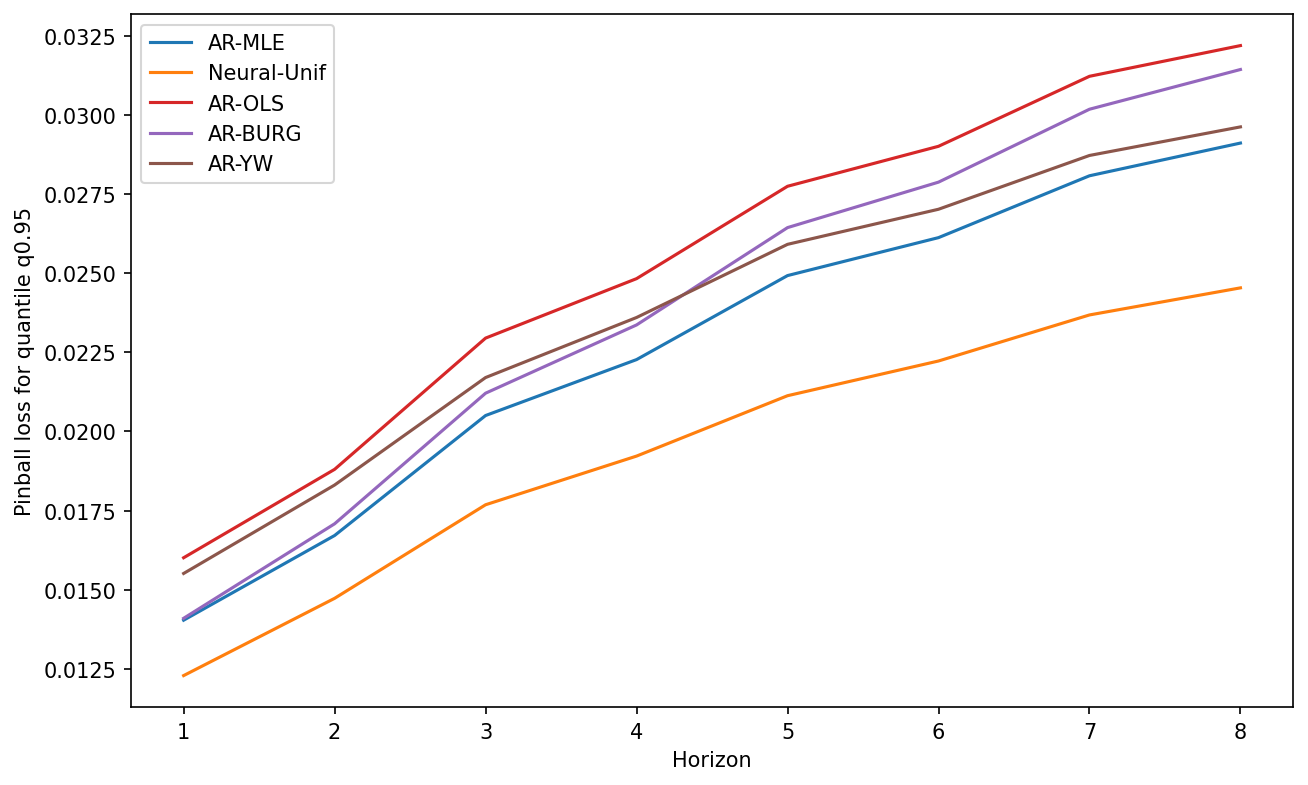}
        \caption{Pinball loss}
        \label{fig:ar_q95_pbl}
    \end{subfigure}
    \hfill
    \begin{subfigure}[t]{0.49\textwidth}
        \centering
        \includegraphics[width=\linewidth]{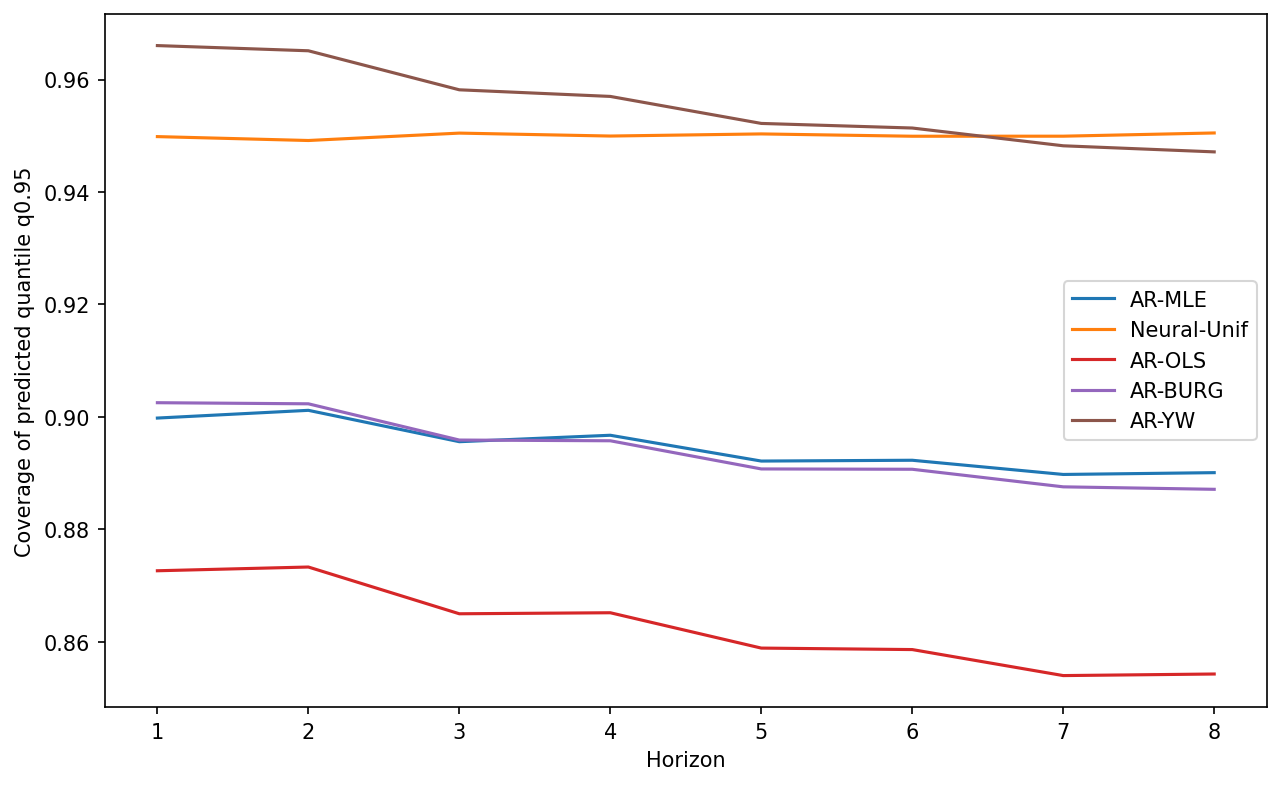}
        \caption{Marginal coverage}
        \label{fig:ar_q95cvr}
    \end{subfigure}

    \vspace{0.4em}

    \begin{subfigure}[t]{0.49\textwidth}
        \centering
        \includegraphics[width=\linewidth]{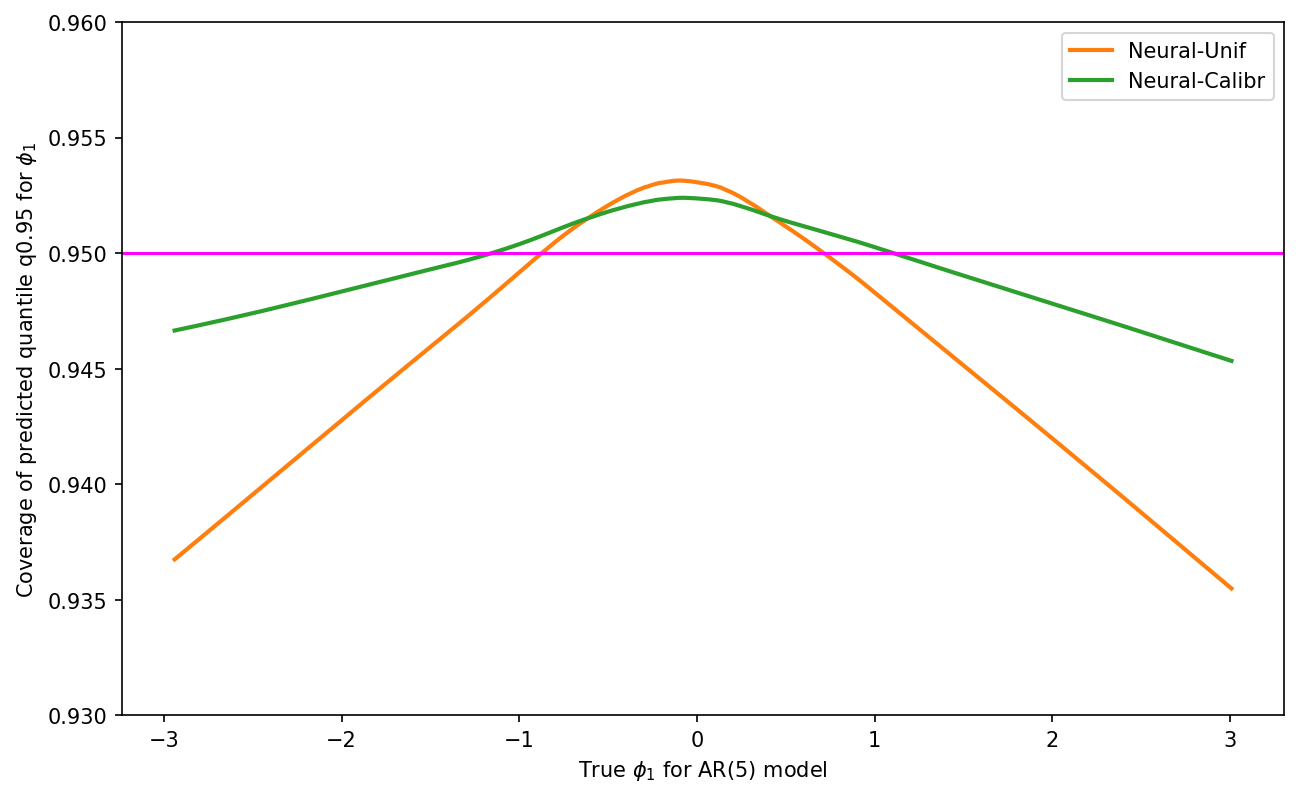}
        \caption{Uniform calibration}
        \label{fig:ar5_phi_cvr_bias}
    \end{subfigure}
    \hfill
    \begin{subfigure}[t]{0.49\textwidth}
        \centering
        \includegraphics[width=\linewidth]{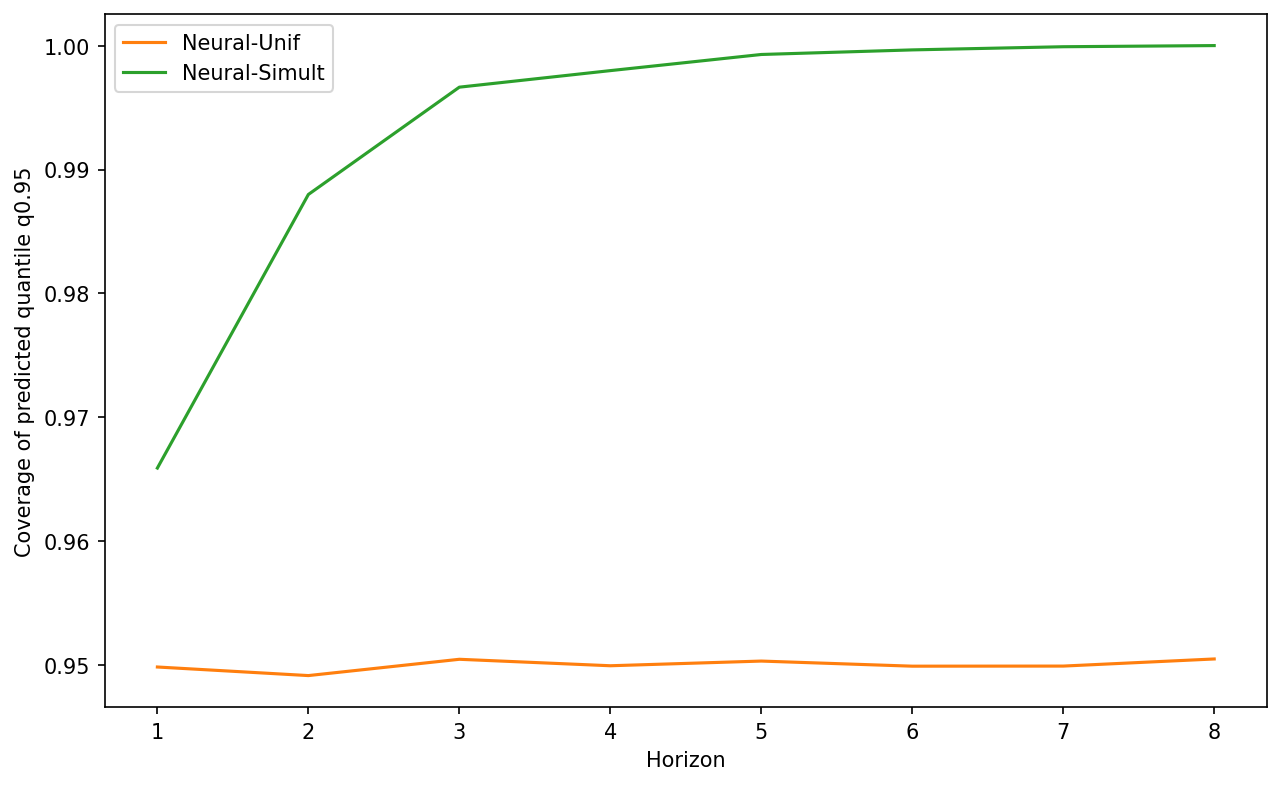}
        \caption{Simultaneous coverage}
        \label{fig:ar5_simult_cvr}
    \end{subfigure}

    \caption{Distributional forecast performance for the 0.95 quantile in AR(5). 
    Panel~(c) is the main result: Neural-Unif achieves nominal coverage only on average, while Neural-Calibr removes the systematic coverage bias across $\phi_1$ and attains near-uniform calibration. 
    Panel~(a) shows that neural estimators also reduce pinball loss; panel~(b) shows that classical intervals undercover because they ignore parameter uncertainty; and panel~(d) shows that Neural-Simult attains nominal simultaneous coverage of the forecast path, unlike pointwise methods.}
    \label{fig:ar5_uncertainty_grid}
\end{figure}

The results in Figure~\ref{fig:ar5_uncertainty_grid} demonstrate how our framework solves the hierarchy of calibration issues:

\begin{enumerate}
    \item \textbf{Forecast Accuracy:} Neural estimators consistently achieve lower Pinball loss than classical estimators across all horizons (Figure~\ref{fig:ar_q95_pbl}).
    
    \item \textbf{Marginal Forecast Calibration:} Classical methods suffer from systematic undercoverage because they ignore parameter uncertainty (Figure~\ref{fig:ar_q95cvr}). In contrast, neural methods achieve nominal marginal coverage.
    
    \item \textbf{Uniform Calibration:} This result parallels the parameter bias result. \textbf{Neural-Unif}, a representative of standard average-risk training (like most TSFMs, GFMs), achieves nominal coverage on average but exhibits systematic coverage bias conditional on $\phi_1$ (a visible ``hump'' in Figure~\ref{fig:ar5_phi_cvr_bias}). \textbf{Neural-Calibr} corrects this, flattening the coverage curve to the nominal level across the parameter space with negligible loss in accuracy.
    
    \item \textbf{Simultaneous Coverage:} A more challenging test of learning the DGP is bounding the entire forecast trajectory $y_{T+1:T+H}$ jointly. Risk management often requires simultaneous bands (e.g., ``the path will stay within these bounds 95\% of the time''). Pointwise methods fail here, achieving only $\sim$75.5\% coverage for a 95\% target due to the compounding of errors over time. \textbf{Neural-Simult} successfully learns the joint distribution, achieving the nominal 95\% simultaneous coverage. Figure~\ref{fig:ar5_simult_cvr} shows the reverse to validate the results: how a simultaneous coverage interval overcovers the pointwise interval as horizon increases.
\end{enumerate}

Taken together, these results show that the proposed framework resolves a central limitation of existing forecasting paradigms:
it can achieve process-level, finite-sample calibration without sacrificing the benefits of shared-data learning.
In the AR$(p)$ class, this yields intervals that are not only accurate on average, but reliable across the parameter space.
Table~\ref{tab:distrib_objectives} summarizes how different approaches target finite-sample distributional objectives.

\begin{table}[htbp]
    \centering
    \caption{Comparison of forecasting approaches for finite-sample distributional objectives}
    \label{tab:distrib_objectives}
    \small
    \begin{tabular}{lcccc}
    \hline
    \textbf{Approach} &
    \textbf{Accuracy} &
    \textbf{Marginal calib.} &
    \textbf{Uniform calib.} &
    \textbf{Simult. coverage} \\
    \hline
    Classical plug-in &
    $\approx$ &
    $\approx$ &
    -- &
    + \\

    Bayesian / PFN / bootstrap &
    $\checkmark$ &
    $\checkmark$ &
    -- &
    + \\

    GFM / TSFM &
    $\checkmark$ &
    $\approx$ &
    -- &
    + \\

    Simulacrum &
    $\checkmark$ &
    $\checkmark$ &
    $\checkmark$ &
    $\checkmark$ \\
    \hline
    \end{tabular}

    \vspace{0.5em}
    \footnotesize
    \textbf{Notes:}
    $\checkmark$ = directly targeted by the standard formulation;
    $\approx$ = approximate, asymptotic, or average-case;
    + = possible with additional modelling, resampling, or calibration;
    -- = not generally targeted.
    Marginal calibration refers to average coverage over the relevant world, prior, empirical distribution, or resampling scheme.
    Uniform calibration refers to coverage conditional on each process or parameter configuration.
    Simultaneous coverage refers to joint coverage of a multi-horizon forecast path.
\end{table}

\section{Application III: Model Selection \& the Forecast Combination Puzzle (Multi-model Worlds)}

\label{sec:modelselpuzzle}
Sections~\ref{sec:ETS} and~\ref{sec:ARp} considered worlds built around a single structural model class.
We now move to \textit{multi-model worlds}, where the world contains several candidate mechanisms and the estimator must learn a higher-level statistical decision.
This setting reflects many practical forecasting pipelines, which select among models for parsimony and interpretation, combine model families to improve accuracy, and hedge against misspecification.

This section considers two such decisions:

\begin{itemize}
    \item \textbf{Model selection.}
    Section~\ref{sec:ETS_modelsel} treats selection among additive ETS subclasses as the target decision, rather than optimizing directly for forecast or parameter-estimation accuracy.
    A neural selector is trained to recover the data-generating subclass and is compared with the standard pipeline of maximum-likelihood estimation followed by AICc-based selection.

    \item \textbf{Forecast combination.}
    Section~\ref{sec:combipuzzle} turns the popular forecast combination heuristic from a post-hoc averaging rule into a direct estimation problem under a combined generative world.
    Rather than fitting ARIMA and ETS separately and then estimating weights for their forecasts, we assume the observed series is generated by a convex combination of ARIMA and ETS components driven by common shocks, defining a novel model class that emerges from combining the two classic models.
    The resulting neural estimator learns the combined-class decision rule directly and achieves state-of-the-art results on many real benchmark datasets.
    It also provides insight into the Forecast Combination Puzzle: even when the estimator is trained to target predictive accuracy directly, its gains over equal weights are modest, explaining why equal-weight combinations remain such a robust benchmark.
\end{itemize}

\subsection{Model Selection for Additive Exponential Smoothing}
\label{sec:ETS_modelsel}

Accurate model selection is essential for parsimony and interpretability. Our goal is to train a \textbf{Neural-Selection} estimator that maximizes the probability of choosing the right subclass from the set of additive ETS models, comparing its accuracy to the standard approach of Maximum Likelihood followed by corrected Akaike Information Criterion (AICc) implemented in \texttt{statsmodels} (\textbf{ETS-AIC}).

Using the simulation scheme from Section~\ref{sec:ETS_def_inference}, we generate synthetic time series across 8 Additive ETS submodels, covering all combinations of Trend (Yes/No), Seasonality (Yes/No), and Seasonal Period (4 or 12).
This ensures coverage of typical Yearly, Quarterly, and Monthly frequencies.
To prevent 'leakage', subclasses have equal probability and series length is controlled so it is not informative of the subclass.
We treat model selection as a supervised classification problem: the neural network is trained to minimize cross-entropy loss, mapping input series to their true model class probabilities.

We evaluate performance on a test set of 128,000 series. Table~\ref{tab:confusion_trend_ets} presents the confusion matrix for Trend detection. \textbf{Neural-Selection} achieves higher accuracy than ETS-AIC. Notably, while ETS-AIC exhibits a strong bias toward selecting complex trended models when none exist (Type I error), the Neural estimator is more robust, showing only a slight conservative bias toward non-trended models.

\begin{table}[htbp]
\centering
\caption{Confusion Matrix for Trend Selection (Additive ETS). \textbf{Neural-Selection (bold)} outperforms \textit{ETS-AIC (italics)}, which exhibits a strong bias toward false positives (detecting trend when none exists).}
\label{tab:confusion_trend_ets}
\begin{tabular}{l c c}
\toprule
& \multicolumn{2}{c}{Predicted Label} \\
\cmidrule(lr){2-3}
True Label & No Trend & Trend \\
\midrule
No Trend & \textbf{0.490} / \textit{0.390} & \textbf{0.012} / \textit{0.113} \\
Trend    & \textbf{0.034} / \textit{0.018} & \textbf{0.463} / \textit{0.479} \\
\bottomrule
\end{tabular}
\end{table}

Regarding Seasonality (Table~\ref{tab:my_confusion_matrix_seasonality_v2}), both methods show a slight bias toward Period=4 (the intermediate complexity between 0 and 12). However, Neural-Selection outperforms ETS-AIC across all categories, achieving an overall accuracy of 97.8\% compared to 94.5\% for AIC.

\begin{table}[htbp]
\centering
\captionsetup{font=footnotesize}
\caption{Confusion Matrix for Seasonality Period Selection. \textbf{Neural-Selection (bold)} achieves 97.8\% overall accuracy, surpassing \textit{ETS-AIC (italics)} at 94.5\%.}
\label{tab:my_confusion_matrix_seasonality_v2}
\begin{tabular}{l c c c}
\toprule
& \multicolumn{3}{c}{Predicted Label} \\
\cmidrule(lr){2-4}
True Label & No Seasonality & Seasonality=4 & Seasonality=12 \\
\midrule
No Seasonality & \textbf{0.321} / \textit{0.296} & \textbf{0.002} / \textit{0.020} & \textbf{0.003} / \textit{0.010} \\
Seasonality=4  & \textbf{0.008} / \textit{0.004} & \textbf{0.332} / \textit{0.325} & \textbf{0.001} / \textit{0.012} \\
Seasonality=12 & \textbf{0.007} / \textit{0.005} & \textbf{0.002} / \textit{0.004} & \textbf{0.325} / \textit{0.324} \\
\bottomrule
\end{tabular}
\end{table}

\begin{figure}[htbp]
    \centering
    \begin{subfigure}[b]{0.48\textwidth}
        \centering
        \includegraphics[width=0.8\textwidth]{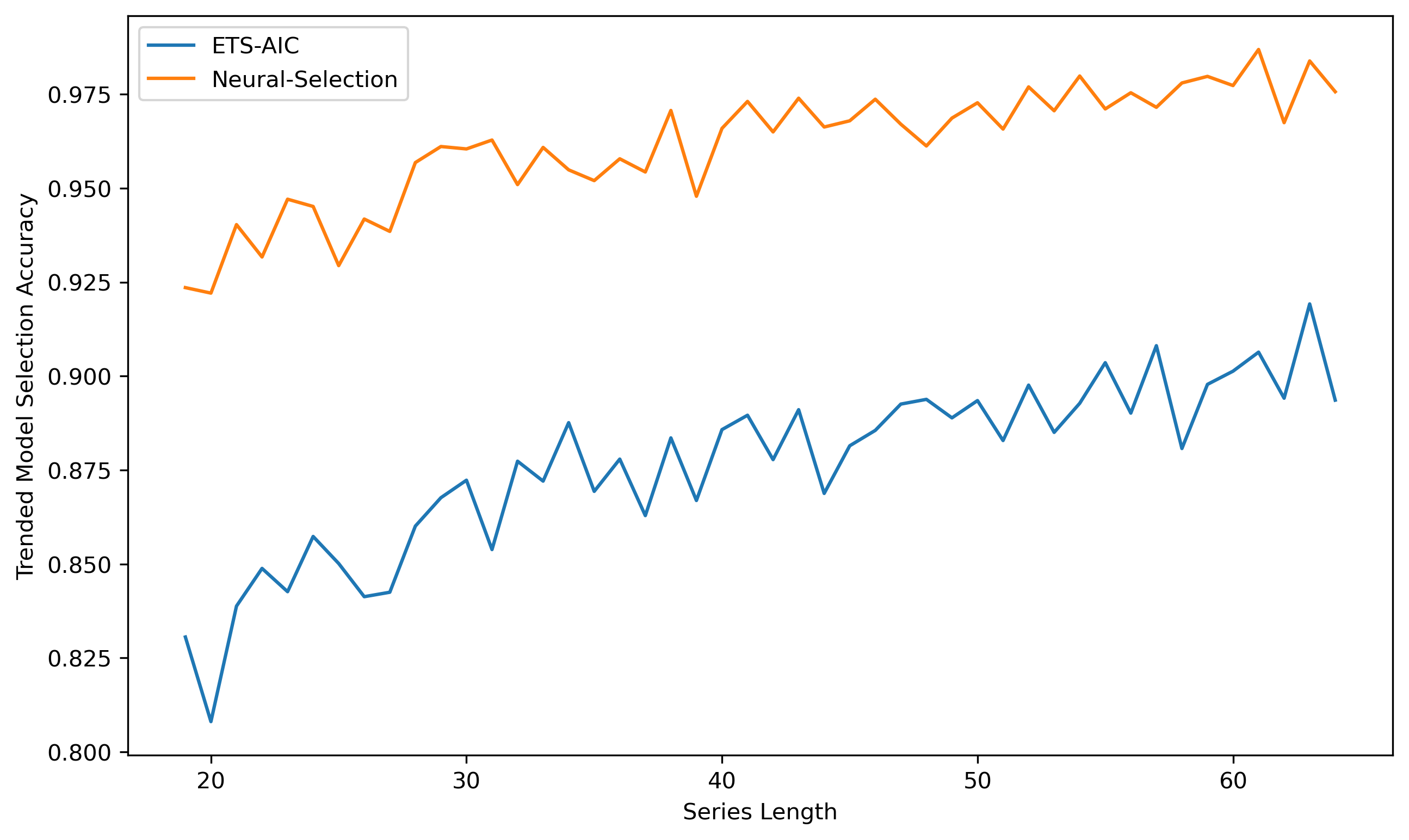}
        \caption{Trend Selection Accuracy}
        \label{fig:ets_aaa_sel_trend}
    \end{subfigure}
    \hfill
    \begin{subfigure}[b]{0.48\textwidth}
        \centering
        \includegraphics[width=0.8\textwidth]{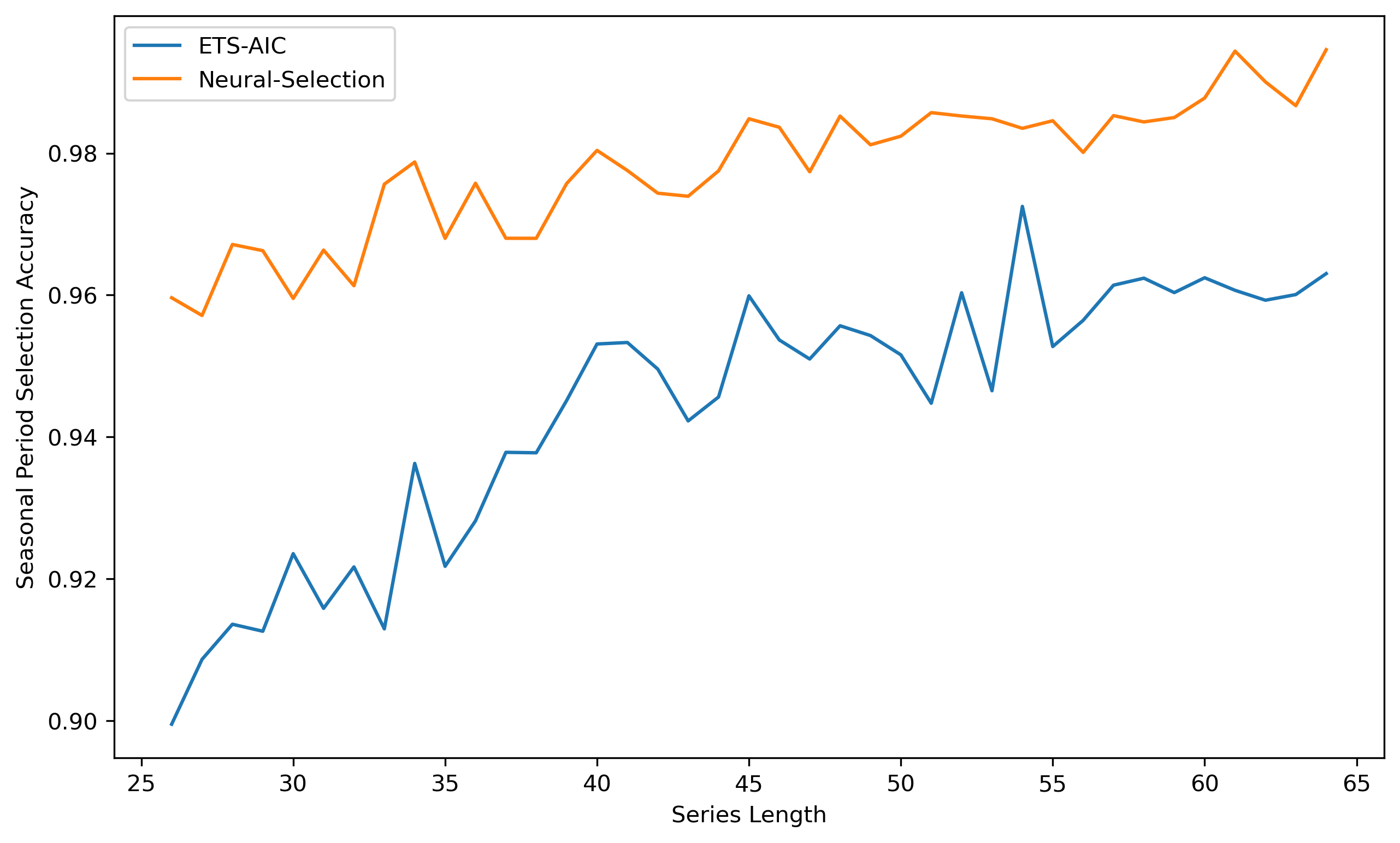}
        \caption{Seasonality Selection Accuracy}
        \label{fig:ets_aaa_sel_season}
    \end{subfigure}
    \caption{Model selection accuracy vs. Series Length. The Neural estimator (Orange) consistently outperforms ETS-AIC (Blue) for both (a) Trend detection and (b) Seasonality period detection.}
    \label{fig:ets_aaa_sel_accu}
\end{figure}

Figure~\ref{fig:ets_aaa_sel_accu} illustrates the scaling of selection accuracy with series length. While both methods improve as the time series length increases, Neural-Selection has better accuracy for lower samples and converges faster toward perfect selection.

\subsection{A Generative Resolution to the Forecast Combination Puzzle}
\label{sec:combipuzzle}

Combining forecasts via weighted averages is a staple in forecasting applications, dominating competitions \cite{makridakis2018m4} and key to success in high-stakes scenarios such as COVID-19 forecasting \cite{sherratt2023predictive}. However, the field faces the ``Forecast Combination Puzzle'': simple equal-weighted combinations often outperform theoretically optimal estimated weights based on historical performance \cite{clemen1989combining, smith2009simple}. This paradoxical effect is attributed to the fact that the error incurred in estimating weights often negates the theoretical gain in predictive power vs a simple `mean of forecasts' baseline\cite{claeskens2016puzzle}.

We propose a resolution to the puzzle by reframing combination not as a post-hoc averaging step, but as a \textbf{generative process}. We define a ``Combined World'' where observations result from a convex combination of two distinct DGPs (e.g., ARIMA and ETS). Importantly, this differs from synthetic data augmentation methods like TSMixup (in Chronos foundation model) \cite{ansari2024chronos}, which combine \textit{independent} series. In the forecast combination problem, candidate models are estimated on the same time series and are thus subject to the same shocks. We capture this dependence by exposing the ARIMA and ETS components to \textit{identical innovations} ($\varepsilon_t$). This creates a ``convex combination with common shocks'' generative model that formalizes the heuristic two-step process. In this combined world, the analytical solution to optimal weight estimation is intractable, but since the ground truth is known, it can be approximated by a neural estimator, \textbf{Neural-Puzzle}. Even though estimation faces identifiability issues (as ARIMA and ETS classes overlap), the Simulacrum successfully recovers an estimator for both optimal weights and optimal forecasts in the sense of squared error directly, without having to estimate underlying component models.

To validate this approach, we trained estimators for the convex combination of the ARIMA and Exponential Smoothing classes. We compare three neural variants:
\begin{enumerate}
    \item \textbf{Pure Baselines:} \texttt{Neural-ARIMA} and \texttt{Neural-ETS}, trained on their respective single-model worlds.
    \item \textbf{Convex Combination (\texttt{Neural-Puzzle}):} Trained on series generated by $y_t = w \cdot y_t^{\text{ETS}} + (1-w) \cdot y_t^{\text{ARIMA}}$, where $w \sim \mathrm{U}(0,1)$. Uniquely, the set of innovations $\varepsilon_t$ is identical for both models.
\end{enumerate}

Simulations for ETS and ARIMA cover Yearly and Quarterly frequencies (Seasonality 0 and 4). ETS follows the sampling described in Section~\ref{sec:ETS_def_inference}.
ARIMA simulations follow an ARIMA$(p,d,q)\times(P,D,Q)_4$ with $p,q \leq 3$, $P,Q \leq 2$, $d + D \leq 2$ 
and drift set to 0 if $d+D = 2$.
This mechanism emulates the heuristically restricted model space in the auto.arima from the \texttt{forecast} R package \cite{hyndman2008automatic}.
AR and MA orders and differencing levels are sampled from the discrete uniform.
AR and MA coefficients follow the uniform PACF inversion (see Section~\ref{sec:ARp}), both for generating stationary AR and invertible MA coefficients. Series lengths are sampled uniformly subject to the constraint that each component model can be estimated given its number of parameters.

We define the generative process for the `Combined World' $y_t$ as follows: 
\begin{enumerate}[nosep]
    \item Draw structure and parameters $\theta_{\text{ETS}}$ and $\theta_{\text{ARIMA}}$ from their respective definitions.
    \item Generate a single sequence of i.i.d. innovations $\varepsilon_{1:T} \sim \mathcal{N}(0,1)$. $T$ is sampled uniformly with maximum of 64. The minimum length is chosen so that both component models can be estimated.
    \item Generate the component series $y^{\text{ETS}}_{1:T}$ and $y^{\text{ARIMA}}_{1:T}$ using their respective recursive equations driven by the \textit{same} $\varepsilon_{1:T}$ sequence.
    \item Normalize both series via min-max normalization.
    \item Construct the final observed series $y_t$ as a weighted sum of the standardized components: 
    $$ y_t = w \cdot y^{\text{ETS}}_t + (1-w) \cdot y^{\text{ARIMA}}_t $$
    where $w \sim \mathrm{U}(0,1)$ represents the mixing weight.
\end{enumerate}

\subsubsection{Real Data: Forecast Accuracy of Puzzle-based Combination}
\label{sec:combipuzzle_results}

Table~\ref{tab:forecast_comparison_cleaned} reports results on the Monash Time Series Archive \cite{godahewa2021monash}. The \textbf{forecast combination} estimator Neural-Puzzle achieves state-of-the-art accuracy, outperforming the pure `Neural-ARIMA' and `Neural-ETS' baselines in almost all datasets and metrics. Critically, it also outperforms the best models in the Monash benchmark, indicated by `*' in the table. \footnote{The benchmark reference values are taken from the Monash Time Series Forecasting Archive comparison available at the time of evaluation, November 2025. The comparison includes classical ARIMA/ETS methods and machine-learning models such as Transformer-, CNN-, N-BEATS-, and CatBoost-based methods.}

\begin{table}[htbp]
\setlength{\tabcolsep}{4pt}
\centering
\caption{Forecasting performance across neural estimators trained on ARIMA, ETS, and combined ARIMA--ETS worlds for yearly and quarterly benchmark datasets.
The ``Convex Win \%'' row reports the proportion of individual series for which \texttt{Neural-Puzzle} achieves lower error than the corresponding single-model estimator.
An asterisk indicates that the reported value improves on the best model in the Monash benchmark comparison snapshot described in the text.}

\label{tab:forecast_comparison_cleaned}
\begin{tabular}{llrrr}
\toprule
\multirow{2}{*}{Dataset} & \multirow{2}{*}{Metric} & \multicolumn{3}{c}{Model} \\
\cmidrule(lr){3-5}
 & & Neural-ARIMA & Neural-ETS & Neural-Puzzle \\ \midrule
\multirow{3}{*}{M4 Yearly} 
 & SMAPE & 14.0 & 13.9 & 13.7* \\
 & MASE & 3.08 & 3.09 & 3.01* \\
 & Convex Win \% & 52.39\% & 54.58\% & --- \\ \midrule

\multirow{3}{*}{M4 Quarterly} 
 & SMAPE & 10.2 & 10.1 & 10.2* \\
 & MASE & 1.15 & 1.15 & 1.13* \\
 & Convex Win \% & 52.54\% & 51.59\% & --- \\ \midrule

\multirow{3}{*}{M3 Yearly} 
 & SMAPE & 15.8 & 16.6 & 16.1 \\
 & MASE & 2.61 & 2.75 & 2.65* \\
 & Convex Win \% & 50.13\% & 54.03\% & --- \\ \midrule

\multirow{3}{*}{M3 Quarterly} 
 & SMAPE & 9.5 & 9.5 & 9.3 \\
 & MASE & 1.11 & 1.14 & 1.11* \\
 & Convex Win \% & 48.93\% & 54.10\% & --- \\ \midrule

\multirow{3}{*}{Tourism Yearly} 
 & SMAPE & 33.2 & 30.6 & 30.0 \\
 & MASE & 3.15 & 2.97 & 2.94 \\
 & Convex Win \% & 62.11\% & 51.98\% & --- \\ \midrule

\multirow{3}{*}{Tourism Quarterly} 
 & SMAPE & 16.1 & 15.0 & 15.0 \\
 & MASE & 1.62 & 1.52 & 1.52 \\
 & Convex Win \% & 57.76\% & 51.32\% & --- \\ \midrule

\multirow{3}{*}{M1 Yearly} 
 & SMAPE & 17.7 & 16.4 & 16.9* \\
 & MASE & 3.67 & 3.31* & 3.45 \\
 & Convex Win \% & 59.02\% & 44.20\% & --- \\ \midrule

\multirow{3}{*}{M1 Quarterly} 
 & SMAPE & 16.6 & 16.0 & 15.6* \\
 & MASE & 1.72 & 1.69 & 1.65* \\
 & Convex Win \% & 53.57\% & 57.94\% & --- \\ \bottomrule
\end{tabular}
\end{table}

\subsubsection{Solving the Puzzle: Comparison with Equal Weights}
\label{sec:combipuzzle_weights}

To assess the forecast-combination puzzle in this setting, Table~\ref{tab:convex_vs_equal} compares \texttt{Neural-Puzzle} with a post-hoc equal-weight average of \texttt{Neural-ETS} and \texttt{Neural-ARIMA}.
The performance gap is narrow, with \texttt{Neural-Puzzle} winning in approximately 49--59\% of cases depending on the dataset.

These results are consistent with the standard explanation of the puzzle: equal weights are a robust heuristic because the gains from unequal weighting are often small relative to the difficulty of estimating weights from finite samples.
Unlike the two-step approaches analyzed in prior literature,  the neural estimator directly targets the combination as the underlying DGP, providing a more complete resolution.

\begin{table}[htbp]
\setlength{\tabcolsep}{4pt}
\centering
\caption{Convex Combination vs. Post-hoc Equal Weights. Values represent the \% of series where the trained Convex estimator yielded lower error than simple averaging. Values near 50\% indicate that the learned combination performs similarly to equal weights, consistent with the robustness of equal weighting in forecast combination.}
\label{tab:convex_vs_equal}
\begin{tabular}{lrr}
\toprule
Dataset & \% Wins (Neural-Puzzle vs Equal Weights) \\ \midrule
M4 Yearly & 50.67\% \\
M4 Quarterly & 50.26\% \\
Tourism Yearly & 59.12\% \\
Tourism Quarterly & 52.87\% \\
M1 Yearly & 48.16\% \\
M1 Quarterly & 54.86\% \\ \bottomrule
\end{tabular}
\end{table}

\section{Related Work}
\label{sec:related-work}
 
We now discuss how these results fit within the broader forecasting and inference literature.

\subsection{Motivating Evidence for Better Estimation of Structural Models as a Way Forward}
\label{subsec:motivation}

Empirical studies on massive datasets suggest that the superior performance of Deep Learning Global Forecasting Models is driven by better generalization rather than the ability to model complex, non-linear patterns \cite{montero2021principles}. This is evidenced by the observation on many benchmark datasets that while neural networks outperform classical models (like ARIMA or ETS) out-of-sample, they often exhibit higher in-sample error, implying that neural networks succeed by avoiding the overfitting common in classical estimation, effectively acting as robust estimators for simple signals. Winner methods in the M4 Competition \citep{smyl2020hybrid} used Exponential Smoothing with a recurrent neural network for the portion of the signal that could not be captured by the structural model, and similar performance was achieved just by combinations of structural models \cite{montero2020fforma}.

Recent literature reinforces the value of simple, structural models when paired with rigorous estimation or preprocessing. \citep{cheng2025arimaplus} demonstrate that a mature preprocessing pipeline allows simple ARIMA models to achieve state-of-the-art performance, matching or outperforming foundation models on the Monash Forecasting Archive. Similarly, \citep{smyl2025local} provides evidence that structural Bayesian exponential smoothing methods can achieve top-tier accuracy. Collectively, these works indicate that the bottleneck in forecasting is often not the \textit{expressiveness} of the model class, but the quality of the \textit{estimator}.

\subsection{Simulation-Based Inference}
\label{sec:simulation-based-approaches}
 
Simulation has long been used in forecasting and inference. \cite{gourieroux1993indirect} introduced \emph{indirect inference}, in which the parameters of a model with an intractable likelihood are recovered by matching an auxiliary statistic computed on observed data to the same statistic computed on simulated data; the efficient method of moments of \cite{gallant1996moments} is a closely related score-based variant. The crucial difference between these approaches and the present framework is amortization. Classical indirect inference solves a separate optimization problem for every dataset, searching over parameters whose simulated auxiliary statistic best matches the observed one. The Simulacrum instead solves the estimation problem once, over an entire generative world, by learning a direct mapping $F$ from observed series to estimands, so that each new series is handled by a single forward pass.
 
This amortization is shared with modern simulation-based inference (SBI), which trains neural networks to approximate inference mappings for models with intractable likelihoods \citep{cranmer2020frontier}. Neural posterior estimation \citep{greenberg2019automatic}, amortized Bayesian inference frameworks such as BayesFlow \citep{radev2020bayesflow}, and Neural Bayes Estimators \citep{SZH24} all learn estimators by simulation. These methods target the Bayesian posterior; in our notation they correspond to the special case in which $\Pi$ is a Bayesian model $\Pi_{\mathrm{Bayesian}}$,
\[
    \widehat F_{\mathrm{NBE}} \;\approx\; \arg\min_F \mathcal{R}(F;\Pi_{\mathrm{Bayesian}}).
\]
Our formulation generalizes this in two respects: the world $\Pi$ need not be a Bayesian model, and the objective need not be posterior recovery but may target conditional, decision-theoretic criteria (Section~\ref{sec:restrictions}). The same amortized, task-distribution structure underlies neural processes \citep{garnelo2018conditional} and meta-learning more broadly \citep{baxter2000model, finn2017model}, with the generative world $\Pi$ playing the role of a task distribution. What distinguishes our setting is the object being optimized: meta-learning typically targets fast adaptation or average accuracy across tasks, whereas we target process-level statistical properties that must hold at each individual task.
 
Another line of research uses simulation to enlarge the data on which a conventional forecaster is trained. Synthetic series have been used to train global models where real data are insufficient \citep{hewamalage2022global}, and synthetic augmentation has been used to improve feature-based forecast model selection \citep{talagala2022fformpp}, often drawing on the GRATIS scheme for generating diverse, realistic series with controllable characteristics \citep{kang2020gratis}. In this literature, the simulation distribution is chosen for diversity or coverage. Our framework generalizes these approaches by proposing deliberate world design paired with a decision-theoretic objective, so that the simulation distribution is no longer a heuristic source of variety but the precise specification of the problem the estimator is required to solve.
 
\subsection{Neural Forecasting Methods}
\label{sec:neural-forecasting-methods}
 
Many recent neural forecasting methods can be written as approximate minimizers of a population risk $\mathcal{R}(F;\Pi)$ for a particular choice of generative world $\Pi$. Making this choice explicit reveals a common structure across methods that otherwise look distinct. 
 
\subsubsection*{Global forecasting models}
 
Global models such as DeepAR \citep{salinas2020deepar}, N-BEATS \citep{oreshkin2020nbeats}, and the Temporal Fusion Transformer \citep{lim2021temporal} train a shared predictor over the empirical distribution of available series,
\[
    \Pi = \Pi_{\mathrm{empirical}}.
\]
This delivers cross-series generalization, but the empirical world is often narrow, so the learner is forced to extrapolate beyond what $\Pi_{\mathrm{empirical}}$ supports. The difficulty is most acute for distributional forecasting, where rare events and tail behavior are scarcely represented in finite corpora. Hybrid worlds allow $\Pi_{\mathrm{empirical}}$ to be expanded explicitly and in a controlled way.
 
\subsubsection*{Time series foundation models}
 
Time series foundation models (TSFMs) such as Chronos \citep{ansari2024chronos} and TimesFM \citep{das2024timesfm} scale training to massive, heterogeneous pretraining worlds $\Pi_{\mathrm{pretrain}}$ assembled from diverse empirical and synthetic series,
\[
    \widehat F_{\mathrm{TSFM}} \approx \arg\min_F \mathcal{R}(F;\Pi_{\mathrm{pretrain}}),
\]
with zero-shot generalization arising when $\Pi_{\mathrm{pretrain}}$ has sufficient support over the test domain. In our framework, TSFMs occupy one end of a spectrum that runs from classical single-mechanism worlds to vast pretraining worlds.
 
\subsubsection*{Synthetically trained forecasters}
 
A distinct line of work trains inference machines on synthetic data alone. In our notation,
\[
    \Pi = \Pi_{\mathrm{synthetic}}, \qquad
    \widehat F_{\mathrm{synth}} \approx \arg\min_F \mathcal{R}(F;\Pi_{\mathrm{synthetic}}).
\]
Prior-data fitted networks \citep{muller2022transformers} learn an inference rule by repeatedly sampling tasks from a synthetic prior and training a network to predict held-out points; the trained network thereby approximates the Bayesian posterior predictive for that prior. TabPFN \citep{hollmann2025accurate} is the tabular instantiation, and ForecastPFN \citep{dooley2023forecastpfn} a variant whose synthetic prior is purpose-built for forecasting. TabPFN-TS \citep{hoo2026tablestotime} shows the premise is general enough that the tabular model itself, pretrained only on synthetic tables, forecasts competitively once timestamps are encoded as tabular features.
 
SarSim0 \citep{oreshkin2026zero} trains standard backbones such as N-BEATS by ordinary risk minimization on an engineered synthetic world of SARIMA processes. In the vocabulary used here, this restricts the world to well-behaved processes and applies designed contamination operators (Section~\ref{sec:robust-design}). On Gift-Eval \citep{aksu2024gifteval}, its estimators outperform the AutoARIMA procedure built on the same SARIMA family, an effect its authors describe as an emergent ``student-beats-teacher'' generalization. The framework developed here clarifies this result. AutoARIMA is a plug-in procedure, and therefore a suboptimal approximation to the Bayes forecaster for the SARIMA world, the rule that integrates over order and parameter uncertainty. A network trained by risk minimization over that world approximates the Bayes forecaster directly.

\subsection{Discussion}
\label{sec:design-space}

Recent progress in neural forecasting has come along two fronts: increasingly expressive architectures, and the enlargement of the world a model is trained on, from a single dataset to large pretraining corpora. Both are, in part, inspired by foundation models for language and vision, where scaling the model and the data has been the dominant lever. One premise of that program does not carry over to time series applications: language and vision are supplied with vast real data, whereas real time series corpora are comparatively scarce, domain-specific, and constrained by licensing. The synthetically trained forecasters of the previous subsection are themselves a response to this, and once that step is taken the generative world becomes a central object of design.

TabPFN \citep{hollmann2025accurate}, a tabular foundation model, shows that a deliberately constructed synthetic prior can equip an estimator with inductive biases that transfer strongly to real tasks. Yet across all of the methods surveyed here, the world $\Pi$ is varied while the objective is held fixed: each is an approximate minimizer of average risk over its world.

The contribution of this paper is to make the estimator itself a first-class object of design. Because the world is simulated, repeated trajectories from any fixed process $\omega$ are available at training time, and the estimator can be trained to target specific statistical properties (Section~\ref{sec:restrictions}). Because the data-generating process is known and can be resampled at will, these properties are verifiable: unbiasedness or coverage at a given $\omega$ can be checked directly by Monte Carlo. A candidate world can then be judged by whether the estimator trained on it meets a stated, in-world property, not only by benchmark accuracy.

This also sharpens a question the framework is equipped to study: whether a general, domain-agnostic prior of the kind TabPFN employs, or a world deliberately matched to a deployment domain, is the better foundation for a given time series task.
 
\section{Conclusions}

This paper introduced the \emph{Simulacrum}, a simulation-based framework for training neural networks as estimators of structural time-series models.
Rather than treating neural networks as generic black-box predictors,
the framework combines a ``generative world'' (i.e. a generalization of data-generating processes), with a decision objective to learn estimators that approximate optimal finite-sample decision rules. This makes it possible to learn estimators for objectives that are central to statistical practice but often difficult to achieve with classical procedures, including low forecast risk, bias reduction, minimax behavior, and process-level calibration. The approach can be used not only to improve estimation within existing model classes, but also to derive estimators for novel classes where even suitable procedures are analytically unavailable or computationally intractable.

Across the applications studied here, the same underlying principle leads to strong and coherent results in both simulated and real-data settings. For exponential smoothing models, neural estimators improve forecast accuracy and can be trained toward lower bias or lower worst-case risk. For the AR$(p)$ class, the framework yields estimators with substantially improved finite-sample behavior, including near-unbiased parameter estimation and near-uniform calibration. For forecast combination, it provides a direct way to learn combination rules in settings where the corresponding optimal estimator is analytically unavailable or impractical to compute, achieving state-of-the-art forecasting accuracy in major benchmarking datasets.

The results also suggest a broader methodological point.
Much of the recent progress in forecasting has been framed as a choice between classical statistical models and highly flexible machine-learning systems, but our findings indicate that this contrast is incomplete.
When estimation is improved sufficiently, simple structural models can remain highly competitive.

More broadly, the paper argues for an additional role for neural methods in time-series analysis. The goal need not be to replace structural models with increasingly flexible forecasting architectures. An alternative is to use neural networks to \emph{solve} estimation problems for structural models that would otherwise be analytically intractable, numerically unstable, or computationally expensive.
In this view, simulation-based pretraining extends rather than abandons the structural modeling tradition.
It preserves parsimony, interpretability, and auditability, while enlarging the set of statistical guarantees that can be pursued in practice.
Some of these guarantees, such as uniform calibration, coverage control, and robustness to worst-case errors, are not only statistically desirable, but operational or regulatory requirements in high-stakes settings.

These capabilities also expand the scientific uses of structural time-series models. When finite-sample bias and miscalibration are reduced, such models become more reliable tools for inference under the umbrella of analysis under temporally correlated error, including applications involving hypothesis testing and uncertainty quantification. More generally, the framework suggests that pretrained estimators could serve as specialized components in larger analytical or agentic workflows, where structural hypotheses are proposed, solved, and evaluated through simulation.

Several directions remain open. One is to expand the class of generative worlds to include richer dynamics, contamination mechanisms, hybrid synthetic--empirical designs, and the study of performance under misspecification of those worlds. Another direction is to broaden the class of decision objectives that can be targeted with the framework. For example, recent work studies estimation problems in which individuals may strategically misreport their data, introducing game-theoretic requirements such as incentive compatibility and robustness to adversarial reporting \cite{haviv2025individual}.


\vskip 0.2in
\bibliographystyle{plain}
\bibliography{mainbiblio}
\appendix

\section{Simultaneous Confidence and Prediction Bands}
\label{app:bands}

Pointwise coverage for neural forecasts and estimates can be targeted with the calibration-adjusted quantile loss.
There are cases when we are interested in coverage along the entire forecast path,
representing the chance for ``at least one'' time-step along the forecast horizon to be outside of the band, for which pointwise intervals can substantially undercover the full forecast path.
Prediction bands are more complex than pointwise, they rely on the joint forecast distribution, which tends to be intractable to model explicitly from data.
In time series, it is very common that samples from the joint distribution of the forecast path are computed by recursively sampling from the more tractable one-step ahead forecast distribution.
This raises an empirical tradeoff: the accumulation of errors in the one-step ahead recursive forecasts vs tractability of direct multi-horizon joint distributions.
The neural network can be trained specifically for this problem.
The forecast path until the desired horizon can be set as the target of the neural network in the training scheme and then bands can be induced by a special loss function. 
The loss function for the bands is computed by minimizing the maximum pinball loss across the forecast horizons (or across the parameter set for estimations).

\[
\mathcal{L}_{\text{band}}^{\tau, \mathcal{H}} = \max_{h \in \mathcal{H}} \mathcal{L}_{\text{quantile}}^{\tau}(h)
\]

Minimizing this loss leads to bands with good simultaneous coverage for the prediction set $\mathcal{H}$, on average across the distribution of the parameter space $\pi(\theta)$.
Confidence bands can be obtained analogously.

\section{Neural Architectures: Invariances and Equivariances}
\label{sec:invariances}

We use neural networks as function approximators; because simulations allow arbitrarily large synthetic datasets, other universal function approximators may work. However, neural architecture matters because it can: (i) enforces desired estimation properties (invariances/equivariances) that are hard to learn by simulation alone; (ii) improves generalization under misspecification via structural bias; and (iii) reduces training time by shrinking the effective hypothesis space.
We focus on four properties and their practical implementations.

\subsection*{Properties}

For the sake of brevity, we present equivariance properties, usually desirable for estimation targets such as forecasts.
Invariances, usually required for parameter estimation, can be induced analogously.

\subsubsection*{Affine Equivariance}
For location/scale transformations $(c,s)$ we target
\begin{equation}\label{eq:affine}
F(s\cdot y_{1:T}+c) = s\cdot F(y_{1:T}) + c,
\end{equation}
which is a known property of many models such as ARIMA or ETS. In simulations, it removes the need to simulate for a vast range of parameters such as the variance of the innovation distribution or initial states.
\textbf{Implementation:} reversible input normalization (e.g., per-series affine rescaling) and output de-normalization, such as min-max normalization.

\subsubsection*{Additive Equivariance}
Useful when there is a known deterministic effect in the model, such as a linear trend, or fixed seasonality,
so the neural estimator may learn all possible trends and seasonality without needing to simulate all such effects explicitly.
\begin{equation}\label{eq:add}
F(y_{1:T}+B\beta)=F(y_{1:T}) + B\beta,
\end{equation}
\textbf{Implementation:} a small regression head estimates $\hat{\beta}$; the network processes residuals $y-B\hat{\beta}$ and re-adds $B\hat{\beta}$ at the output.\\

\subsubsection*{Time-Translation Equivariance}
For the shift operator $(\mathcal{T}_\Delta y)_t=y_{t-\Delta}$, useful in finite order models such as AR($p$) or more general models with finite context. Under misspecification, it can impose the guarantee that the output is a function of limited recent observations.
\begin{equation}\label{eq:shift}
F(\mathcal{T}_\Delta y_{1:T})=\mathcal{T}_\Delta F(y_{1:T}),
\end{equation}
\textbf{Implementation:} causal 1D CNNs with kernel size matching the desired context length in the output layer.

\subsubsection*{Transformation Equivariance}
Useful for processes that become approximately additive/Gaussian after a transformation (e.g. Box-Cox or Yeo–Johnson transformations). We target
\begin{equation}\label{eq:transform}
F\!\left(g_\lambda(y)\right)=g_\lambda\!\left(F(y)\right),
\end{equation}
where $g_\lambda$ is a parametric, invertible transform.\\
\textbf{Implementation:} a learnable, invertible head applies $g_\lambda$ to inputs and $g_\lambda^{-1}$ to outputs.

\subsection*{Architecture used in the experiments}
All experiments in this work use a \emph{feedforward} DenseNet~\cite{huang2017densely} (no convolutions). While the framework supports the various inductive biases described above, in these experiments we strictly enforce only \textbf{location and scale equivariance} (unless stated otherwise). This simplifies the performance comparison; the implementation of other architectural constraints (such as those specific to certain processes) is left for future work.

We apply reversible per-series min-max normalization for location/scale handling, and we support variable-length inputs by:
\begin{itemize}
    \item padding shorter series to a fixed window (e.g., $T_{\max}=64$);
    \item encoding the padded values with the special value -1, distinct from the [0,1] range induced by min-max normalization.
\end{itemize}

\section{Additional Experiments Exponential Smoothing}

\subsection{Inference for Simple Exponential Smoothing}

Comparison on inference performance, MSE and $\text{Bias}^2$, averaging across $\alpha$ is seen in Table~\ref{tab:ses_inference}.
The top performing method for MSE is the Neural-Unif that precisely tries to optimize this quantity.
The Neural-Debias estimator has strongly reduced the bias at a MSE cost compared to Neural-Unif, as expected from Figure~\ref{fig:ses_alpha_side_by_side}(right panel).
Maximum Likelihood is dominated overall for MSE and $\text{bias}^2$ by all neural estimators at this sample size. The Neural-Minimax estimator sits in an interesting position in the bias-variance tradeoff compared to the top performer Neural-Unif: in terms of Root MSE, it is only 5\% worse but it is 45\% better in terms of Root $\text{bias}^2$.

\begin{table}[htbp] 
\setlength{\tabcolsep}{4pt} 
\centering 
\caption{Inference errors for $\alpha$ in the SES model. Neural-Unif outperforms the equivalent Bayesian method. Neural-Debias largely reduced bias. Neural Minimax sits at a good bias-variance tradeoff.} 
\label{tab:ses_inference} 
\begin{tabular}{llrrrrr} 
\toprule
\multirow{2}{*}{Parameter} & \multirow{2}{*}{Error Type} & \multicolumn{5}{c}{Model} \\ 
\cmidrule(lr){3-7} 
                   &                             & SES-MLE & Neural-Unif & Neural-Debias & Neural-Minimax & Bayes-Unif\\ 
\midrule
\multirow{2}{*}{Alpha ($\alpha$)} & MSE             & 0.093 & 0.037 & 0.053 &  0.038 & 0.039\\
                   & $\text{bias}^2$    & 0.014 & 0.013 & 0.001 & 0.006 & 0.019\\
\bottomrule
\end{tabular}
\end{table}

\begin{figure}[b]
        \centering
        \includegraphics[width=0.9\linewidth]{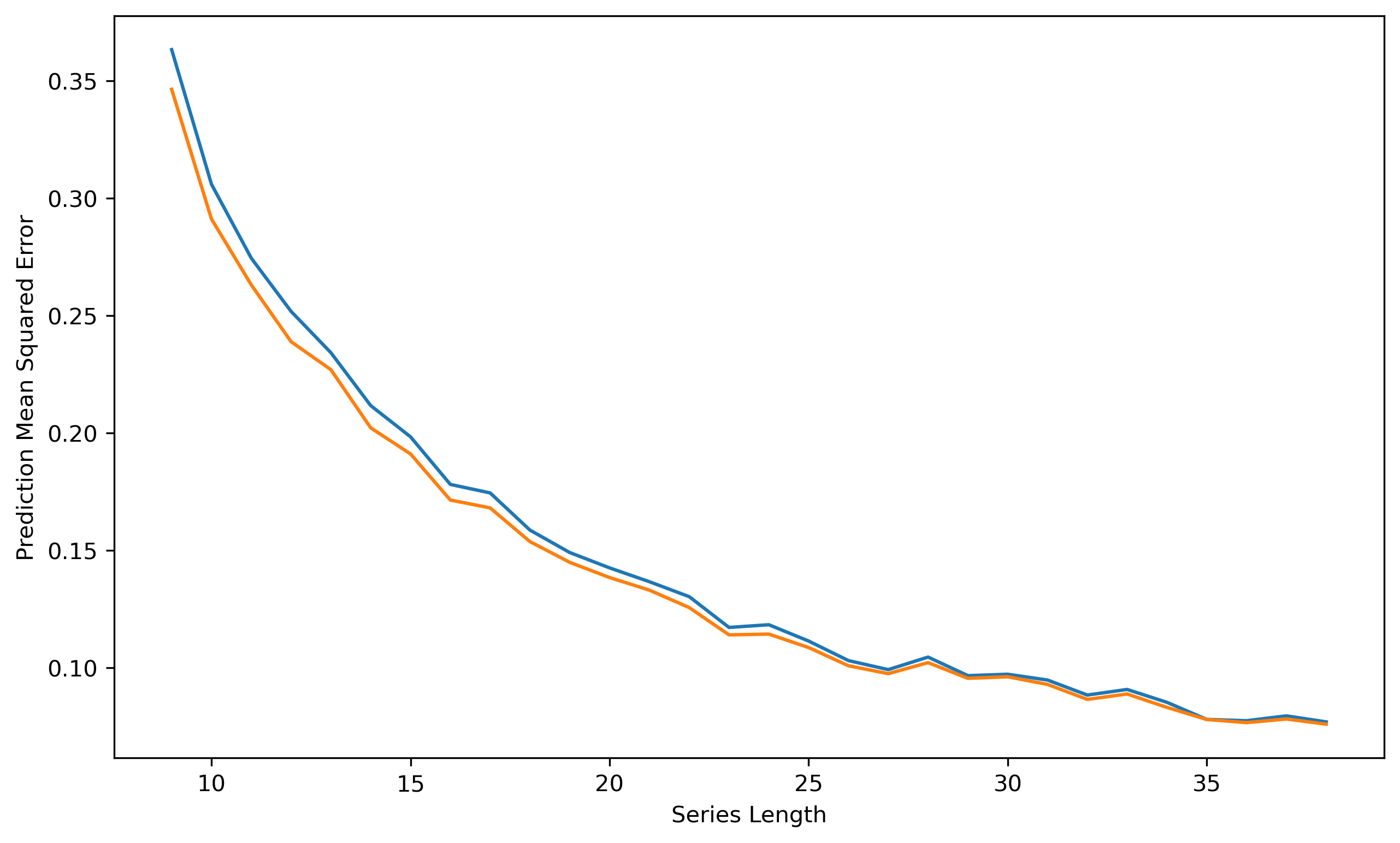} 
        \caption{Squared Prediction Error of SES-MLE (blue) and Neural estimator (orange) vs Time Series Length.}
        \label{fig:ses_length_mse}
\end{figure}

\section{Empirical accuracy of Neural-ETS in M1 competition dataset}

\begin{table}[htbp] 
\centering
\caption{Average MAPE: all M1 Competition data (1001). The Neural-ETS estimator gets a strong performance compared to other methods that participated in the M1 Competition.}
\label{tab:avg_mape_all_data_no_avg_fh}
\resizebox{\textwidth}{!}{
\begin{tabular}{@{}l rrrrrrrrrr r@{}} 
\toprule
& \multicolumn{10}{c}{Forecasting Horizons} & \\
\cmidrule(lr){2-11} 
METHODS & \multicolumn{1}{c}{1} & \multicolumn{1}{c}{2} & \multicolumn{1}{c}{3} & \multicolumn{1}{c}{4} & \multicolumn{1}{c}{5} & \multicolumn{1}{c}{6} & \multicolumn{1}{c}{8} & \multicolumn{1}{c}{12} & \multicolumn{1}{c}{15} & \multicolumn{1}{c}{18} & \multicolumn{1}{c}{n(max)} \\
\midrule
NAIVE 1      & 11.9 & 16.9 & 17.0 & 17.7 & 22.4 & 25.0 & 24.7 & 17.1 & 24.9 & 24.9 & 1001 \\
Mov.Averag   & 13.0 & 18.1 & 17.6 & 18.4 & 22.8 & 25.0 & 23.2 & 17.8 & 23.3 & 23.3 & 1001 \\
Single EXP   & 11.2 & 16.5 & 16.2 & 17.1 & 21.6 & 24.1 & 22.6 & 17.5 & 23.3 & 23.3 & 998 \\
ARR EXP      & 12.4 & 18.9 & 17.6 & 19.0 & 21.7 & 24.6 & 21.7 & 18.7 & 24.2 & 24.2 & 1001 \\
Holt EXP     & 11.4 & 15.9 & 16.6 & 18.6 & 23.7 & 26.8 & 30.1 & 26.5 & 39.5 & 39.5 & 998 \\
Brown EXP    & 11.4 & 15.9 & 17.6 & 18.8 & 24.2 & 27.8 & 31.4 & 29.3 & 43.9 & 43.9 & 1001 \\
Quad.EXP     & 12.8 & 18.0 & 21.1 & 23.1 & 31.9 & 38.3 & 52.8 & 56.0 & 91.3 & 91.3 & 1000 \\
Regression   & 17.8 & 21.8 & 22.1 & 21.0 & 25.8 & 26.7 & 28.0 & 30.8 & 51.9 & 51.9 & 997 \\
NAIVE2       & 9.6  & 11.3 & 13.3 & 14.6 & 18.4 & 19.9 & 19.1 & 17.1 & 21.9 & 21.9 & 1001 \\
D Mov.Avrg   & 11.5 & 14.9 & 17.0 & 17.8 & 21.5 & 22.3 & 20.6 & 17.8 & 23.2 & 23.2 & 1001 \\
D Sing EXP   & 8.0  & 11.6 & 13.2 & 14.1 & 17.7 & 19.5 & 17.9 & 16.9 & 21.1 & 21.1 & 998 \\
D ARR EXP    & 9.4  & 13.5 & 14.0 & 15.3 & 18.1 & 20.2 & 18.0 & 17.1 & 21.4 & 21.4 & 1001 \\
D Holt EXP   & 8.7  & 11.0 & 13.3 & 15.2 & 19.1 & 21.6 & 24.8 & 23.9 & 33.7 & 33.7 & 998 \\
D BrownEXP   & 8.7  & 10.9 & 13.8 & 15.0 & 19.7 & 21.1 & 24.5 & 23.1 & 30.8 & 30.8 & 1001 \\
D Quad.EXP   & 9.8  & 12.7 & 16.6 & 18.8 & 25.7 & 31.0 & 45.1 & 40.7 & 64.4 & 64.4 & 1001 \\
D Regress    & 15.6 & 16.9 & 19.1 & 18.3 & 21.9 & 23.0 & 24.2 & 29.7 & 49.1 & 49.1 & 997 \\
WINTERS      & 8.7  & 10.9 & 13.2 & 14.9 & 19.0 & 21.5 & 24.3 & 23.0 & 32.8 & 32.8 & 998 \\
Autom. AEP   & 9.1  & 11.9 & 13.4 & 13.7 & 17.9 & 20.3 & 20.3 & 19.3 & 24.8 & 24.8 & 1001 \\
Bayesian F   & 11.2 & 12.8 & 14.5 & 15.2 & 19.8 & 22.3 & 22.6 & 18.9 & 23.5 & 23.5 & 997 \\
Combining A  & 8.1  & 10.4 & 12.1 & 13.3 & 16.7 & 19.2 & 19.7 & 18.6 & 24.2 & 24.2 & 1001 \\
Combining R  & 8.5  & 11.1 & 12.8 & 13.8 & 17.6 & 19.2 & 18.9 & 18.4 & 23.3 & 23.3 & 1001 \\
\midrule
Chronos-2    & 9.1  & 12.3 & 13.4 & 14.3 & 17.9 & 21.4 & 18.2 & 18.3 & 23.9 & 40.4 & 1001\\
\midrule
Neural-ETS    & 8.4 & 11.1& 12.3 & 13.6 & 16.8 & 19.1 & 18.7 & 16.9 & 21.4 & 25.8 & 998 \\
\bottomrule
\end{tabular}%
}
\end{table}

\end{document}